\documentclass[a4paper,fleqn,]{cas-dc}

\usepackage[USenglish]{babel}
\usepackage[authoryear,longnamesfirst]{natbib}
\usepackage{microtype}

\ExplSyntaxOn
\cs_set:Npn \__first_footerline: {  }               %
\RenewDocumentCommand \printorcid {  } { }          %
\cs_set:Npn \__first_foot: { \parbox[t]{\textwidth}{ \rule{\textwidth}{.2pt}\\ \centering\sffamily\small \thepage } }
\cs_set:Npn \__cas_foot:   { \parbox[t]{\textwidth}{ \rule{\textwidth}{.2pt}\\ \centering\sffamily\small \thepage } }
\ExplSyntaxOff
\newcommand{\VECTOR}{\fbox{\phantom{\textsf{I}}}}
\newcommand{\TENSOR}{\fbox{\phantom{X}}}

\usepackage{arydshln}
\usepackage{bm}
\usepackage[inkscapearea=page]{svg}
\graphicspath{{figures}}
\svgpath{{figures}}

\usepackage[normalem]{ulem}
\usepackage{float}

\usepackage[OT2,OT1]{fontenc}
\newcommand\cyr
{\renewcommand\rmdefault{wncyr}
	\renewcommand\sfdefault{wncyss}
	\renewcommand\encodingdefault{OT2}
	\normalfont
	\selectfont}
\DeclareTextFontCommand{\textcyr}{\cyr}

\def\tsc#1{\csdef{#1}{\textsc{\lowercase{#1}}\xspace}}
\tsc{WGM}
\tsc{QE}

\begin{document}
\let\WriteBookmarks\relax
\def\floatpagepagefraction{1}
\def\textpagefraction{.001}

\shorttitle{}    

\shortauthors{}  

\title [mode = title]{Modeling, planning and control of robotic unicycles}

\author[1]{M\'{a}t\'{e} B. Vizi}[
    orcid=0000-0001-7372-1296
]

\cormark[1]

\ead{vizi@mm.bme.hu}

\affiliation[1]{organization={Department of Applied Mechanics, 
		Budapest University of Technology and Economics},
            city={Budapest},
            country={Hungary}}

\author[1,2]{D\'{e}nes T\'{a}k\'{a}cs}

\affiliation[2]{
	organization={HUN-REN--BME Dynamics of Machines Research Group},
    city={Budapest},
    country={Hungary}
}

\author[1,2]{G\'{a}bor St\'{e}p\'{a}n}

\author[3,4,5]{G\'{a}bor Orosz}%

\affiliation[3]{
	organization={Department of Mechanical Engineering, University of Michigan},
	city={Ann Arbor},
	postcode={48109}, 
	state={MI},
	country={USA}}
\affiliation[4]{
	organization={Department of Civil and Environmental Engineering, University of Michigan},
	city={Ann Arbor},
	postcode={48109}, 
	state={MI},
	country={USA}}
\affiliation[5]{
	organization={University of Michigan Transportation Research Institute},
	city={Ann Arbor},
	postcode={48109}, 
	state={MI},
	country={USA}}

\cortext[1]{Corresponding author}

\begin{abstract}
This article presents an integrated path-planning and control framework tailored to the unique requirements of robotic unicycles.   
	A two-actuator unicycle design is introduced to execute a variety of maneuvers.
	The proposed path-planning method decomposes the
	path into straight and curved segments, dedicated to acceleration and turning maneuvers,
	respectively. The curvature profiles of the curved segments are optimized while accounting for closed-loop control performance and wheel-slip constraints.
The effectiveness of the proposed integrated approach is demonstrated through numerical simulations.
\end{abstract}


\maketitle

\section{Introduction}

Electric unicycles are an emerging class of micromobility vehicles that are rapidly spreading in urban environments for commuting and last-mile transportation. 
Their appeal stems from a distinctive combination of properties: they are agile and nimble, enabling tight turns and rapid maneuvering; lightweight and compact, making them easy to carry, store, and deploy; and energy efficient, supporting practical daily use. 
These advantages motivate robotic unicycle platforms, but their inherent instability and nonholonomic rolling constraints make modeling, planning, and control especially challenging.

The first conceptual robotic unicycle design known to the authors appeared in~\citep{Ozaka_1980_stability}, using a point mass and pendulum actuator somewhat similar to the approach considered in this paper. 
Since then, several designs with different actuation mechanisms have been proposed. Most systems use two or three actuators for propulsion and maneuvering, including pendulums~\citep{Schoonwinkel_1987,Zenkov_Bloch_Marsden_2002}, flywheels~\citep{Schoonwinkel_1987,vos_VonFlotow_1990,cao2023autonomous}, and robotic arms~\citep{gim2024ringbot}. 
Gyroscopes are also commonly used for lateral stabilization in unicycles and bicycles~\citep{Brown_1996_Single,han2022coordinated,wang2023gyroscopic}. 
Humanoid-type unicycles have also been investigated~\citep{Suzuki_Moromugi_Okura_2014}. 
Key challenges include accurate wheel-ground contact modeling and path-following control design~\citep{sharp2010stability}.

In addition to robotic unicycles, the robotics community has investigated a wide range of wheeled mobile robot designs that use rolling as the primary locomotion principle. 
Examples include wheel-transforming robots~\citep{kim2014wheeltrans,Zheng2023wheeltrans,Ju2025wheeltrans}, mecanum-wheel robots~\citep{Indiveri2009mecanuum,Cheah2023mecanuum}, snake-like robots~\citep{Fukuoka2023snake,Cheah2023mecanuum}, reconfigurable robots~\citep{Cheah2023mecanuum,Qi2024wheelbase}, in-pipe robots~\citep{segon2005pipe,Kwon2012pipe}, and wheeled bipedal and quadrupedal robots such as the Unitree~B2-W and Lynx~M20.

These systems demonstrate the broader interest in rolling locomotion, but physical unicycles pose a distinct challenge because balance, steering, and wheel–ground contact are tightly coupled in three dimensions.
We emphasize that this work concerns the dynamic model of a physical unicycle, rather than the planar kinematic unicycle model commonly used for mobile-robot path planning.
Existing studies have not yet provided a unified framework that integrates three-dimensional dynamic modeling, path planning, feedback control, and wheel-ground contact-force analysis for a highly unstable physical unicycle. 
Such integration is essential because feasible paths and stabilizing controllers are tightly coupled through the nonholonomic rolling constraints and the balancing dynamics.

	Our prior work \citep{vizi2025tcst} introduced a unicycle model with actuation to control the lateral dynamics. Using the straight rolling steady state as a reference, linear feedback was developed to execute simple maneuvers.
	In \citep{vizi2024mecc}, a path-following formulation was added based on \citep{qin2022nonholonomic}, making the lateral and heading errors that could be used directly for feedback control.
	In \citep{vizi2025acc} we improved the mechanical model by introducing actuation for the longitudinal dynamics. It was shown that near the straight rolling steady state, the model can be decomposed into lateral and longitudinal subsystems.
\newpage

	The main contributions of this paper are threefold.
	First, in addition to deriving the equations of motion, the wheel-ground contact force components are calculated in closed form which enables us to 
	monitor the friction demand during the motion.
	Second, the path-based coordinate transformation and a clothoid-based path-planning method are discussed in detail, and their integration with the controller is demonstrated.
	Third, it is shown that this unicycle model can exhibit shimmy-like loss of stability without accounting for tire elasticity. 

The rest of the paper is organized as follows. 
Section~\ref{sec:mechmodel} introduces the robotic unicycle design and derives the corresponding governing equations. 
Section~\ref{sec:controlDesign} presents the motion-planning framework and control design, while Section~\ref{sec:simulation} optimizes and evaluates the design through numerical simulations. 
Section~\ref{sec:conclusion} concludes the paper and outlines future research directions.

\section{Modeling Unicycle Dynamics}\label{sec:mechmodel}

When modeling unicycle dynamics, we assume that the wheel rolls without slipping, resulting in a nonholonomic mechanical system \citep{Voronets1901,Hamel1938,Kane1961,Gantmacher_1970,NeiFuf1972,Koon_Marsden_1997,OstAng1998,Bloch_2003}. 
The governing nonlinear differential equations are derived in closed form using the Appellian approach \citep{Appell_1900,Gibbs_1879,qin2022nonholonomic}. 
This representation yields fewer equations than the Lagrangian formalism because the rolling-wheel kinematic constraints are eliminated. 

\begin{figure}[!t]
	\centering
	\includegraphics[scale=1.0]{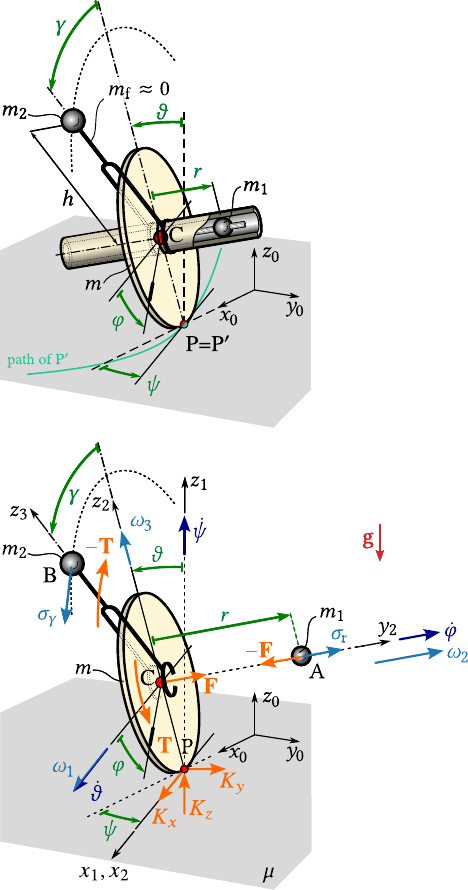}%
	\caption{(Top) Mechanical model of the robotic unicycle. (Bottom) Coordinate frames, kinematic variables and forces.%
	}
	\label{fig:motivation}%
\end{figure}

Our robotic unicycle consists of three bodies: the wheel, the inverted pendulum, and the point mass, which is moved along the axle, see Figure~\ref{fig:motivation}.
One actuator, an electric motor, exerts torque $T$ between the wheel and the inverted pendulum. 
This is responsible for driving the vehicle forward while simultaneously balancing the pendulum.
The other actuator, a linear motor, exerts force $F$ between the wheel and the point mass. 
This is responsible for turning the wheel as well as for lateral balancing.

The wheel is modeled as a thin disc with mass~$m$ and radius~$R$.
The position of the wheel center $\mathrm C$ is given by the vector
\begin{equation} \label{eq:positionG}
	\mathbf r_{\mathrm C} = %
    \begin{bmatrix}
	x_{\mathrm C}
	\\ y_{\mathrm C}
	\\ z_{\mathrm C}
    \end{bmatrix}_{{\mathcal F}_{0}},
\end{equation}
where the subscript indicates that the vector components are resolved in the Earth-fixed coordinate frame $\mathcal F_0$.
The orientation of the wheel is described using the yaw, tilt, and pitch angles, $\psi$, $\vartheta$, and $\varphi$, respectively, see Figure~\ref{fig:motivation}.
With the help of these angles, three additional coordinate frames are introduced.
Frame~$\mathcal F_1$ is obtained by rotating frame~$\mathcal F_0$ with the yaw angle~$\psi$ around the $z_0$ vertical axis and moving the origin to the wheel-ground contact point~$\mathrm P $.
Frame~$\mathcal F_2$ is obtained by rotating frame~$\mathcal F_1$ with the tilt angle~$\vartheta$ around the $x_1$~axis and shifting the origin to the wheel center~$\mathrm C $.
Frame~$\mathcal F_3$ is obtained by rotating frame~$\mathcal F_2$ with the pendulum angle~$\gamma$ around the $y_2$~axis, while keeping the origin at the wheel center~$\mathrm C$.

The wheel is assumed to roll on the horizontal ground.
At a given time instance, point $\mathrm P$ of the wheel is in contact with the ground.
The kinematic condition of rolling is 
\begin{equation}\label{eq:vP}
	\mathbf v_{\mathrm P} = \mathbf 0\,,%
\end{equation}
that is, point $ \mathrm P$ is the velocity pole of the wheel (also called the instantaneous center of velocities).
The velocity $\mathbf v_{\mathrm P}$ can be calculated with the transport formula
\begin{equation}\label{eq:vP_transportFormula}%
	\mathbf v_{\mathrm P} = \mathbf v_{\mathrm C} + \bm \omega \times \mathbf r_{\mathrm{CP}}\,,
\end{equation}
based on the velocity $\mathbf v_{\mathrm C}$ of the wheel center $\mathrm C$, the angular velocity $\bm \omega$ of the wheel, and the position vector $\mathbf r_{\mathrm{CP}}$ pointing from $\mathrm C $ to $\mathrm P$. 
These can be expressed as:
\begin{equation}\setlength\arraycolsep{1ex} \label{eq:vG_velocity}
		\mathbf v_{\mathrm C} 
        \!=\!  
        \dot {\mathbf r}_{\mathrm C} 
        \!=\!
		\begin{bmatrix}
		\dot x_{\mathrm C} \\
		\dot y_{\mathrm C} \\
		\dot z_{\mathrm C}
		\end{bmatrix}_{{\mathcal F}_{0}}\!\!,
        \quad
		\bm \omega 
        \!=\textbf{} 
		\begin{bmatrix}
			\dot\vartheta \\
			\dot\varphi + \dot\psi\sin\vartheta \\
			\dot\psi\cos\vartheta 
		\end{bmatrix}_{{\mathcal F}_{2}}\!\!,
		\quad
		\mathbf r_{\mathrm{CP}} 
        \!=\textbf{} 
		\begin{bmatrix}
			0 \\
			0 \\
			-R 
		\end{bmatrix}_{{\mathcal F}_{2}}\!\!.
\end{equation}
	
	Transforming all vector to $\mathcal F_2$, equations~\eqref{eq:vP}--\eqref{eq:vG_velocity} yield two kinematic constraints
	\begin{equation}\label{eq:kinemConstraints}
		\begin{split}
			\dot x_{\rm C} &= \dot \psi R \cos{\psi} \sin{\vartheta} + \dot \vartheta R \sin{\psi} \cos{\vartheta} +  \dot \varphi R \cos{\psi}\,, 
			\\
			\dot y_{\rm C} &= \dot \psi R \sin{\psi} \sin{\vartheta}  - \dot \vartheta R \cos{\psi} \cos{\vartheta} + \dot \varphi R \sin{\psi}\,, 
		\end{split}
	\end{equation}
	and one geometric constraint
	\begin{equation}\label{eq:geom_constr_eq_z}
		\dot z_{\mathrm C} = - \dot \vartheta  R \sin{\vartheta}\, \quad\rightarrow\quad  z_{\mathrm C} = R \cos{\vartheta}\,.
	\end{equation}
	Geometric constraints are also called holonomic constraints, these only depend on positions and angles.
	Kinematic constraints are also called nonholonomic constraints, these depend on velocities also and they cannot be integrated.

	The point mass~$m_1$ is marked with point~$\mathrm A$ and it can only move along the axle, i.e., along the the $y_2$ axis, that is, its position is given by
	\begin{equation}
		\mathbf r_{\mathrm{CA}} = 
        \begin{bmatrix}
		0 \\ r \\ 0
        \end{bmatrix}_{{\mathcal F}_{2}},
	\end{equation}
	where $r$ is the signed distance of points $\mathrm C$ and $\mathrm A$.
	The internal force
	\begin{equation}
		\mathbf F = 
		\begin{bmatrix}
        0 \\ F \\ 0
        \end{bmatrix}_{{\mathcal F}_{2}},
	\end{equation}
	acts on the wheel, while $-\mathbf F$ acts on the point mass.

	The pendulum is assumed to be a `mathematical' pendulum consisting of  the point mass $m_2$ located at point $\mathrm B$ and a massless fork that is attached to the axle such that the pendulum rotates around the $y_2$ axis with angle $\gamma$ and the mass' position is given by 
	\begin{equation}
		\mathbf r_{\mathrm{CB}} = 
        \begin{bmatrix}
		0 \\ 0 \\ h
        \end{bmatrix}_{{\mathcal F}_{3}}.
	\end{equation}
	The internal torque
	\begin{equation}
		\mathbf T = %
		\begin{bmatrix}
        0 \\ T \\ 0
        \end{bmatrix}_{{\mathcal F}_{2}},
	\end{equation}
	acts on the wheel, while $-\mathbf T$ acts on the pendulum.

	The unicycle system would have ${N=6+3+3=12}$ degrees of freedom without considering the constraints.
	Constraint \eqref{eq:geom_constr_eq_z} and the constrained motions of the masses $m_1$ and $m_2$ together represents ${n_{\mathrm{g}} = 1+2+2=5}$ geometric constraints. 
	Furthermore, there are ${n_{\mathrm{k}}=2}$ kinematic constraints \eqref{eq:kinemConstraints}; so the unicycle is a nonholonomic mechanical system with ${N-n_{\mathrm{g}}-n_{\mathrm{k}}/2 = 6}$ degrees of freedom.
	That is, the equations of motion consist of ${2(N-n_{\mathrm{g}})-n_{\mathrm{k}} = 12}$ first-order differential equations
    out of which ${N-n_{\mathrm{g}}=7}$ are written for the generalized coordinates and ${N-n_{\mathrm{g}}-n_{\mathrm{k}}=5}$ are written for the velocity coordinates.

	A common method to derive the equations of motion of a nonholonomic mechanical system is the generalized Lagrangian equations, which is also called the Routh--Voss equations~\citep{Routh_1884,Voss_1885}.
	This method extends the well-known Lagrangian equations of the second kind (originally derived for holonomic systems) and yields a set of differential algebraic equations.
	A lesser known method is the Appellian approach~\citep{Appell_1900, Gibbs_1879, qin2022nonholonomic} that yields a set of first-order ordinary differential equations as the most compact representation of the nonholonomic mechanical system.
	The downside of the Appellian approach is that it requires the calculation of accelerations and the so-called acceleration energy during the intermediate steps of the derivation. 
	As demonstrated by the authors' earlier works \citep{vizi2025tcst,vizi2024mecc,cao2023autonomous} for unicycle dynamics, it is worth taking the Appellian approach, as it significantly simplifies the dynamical analysis and the control design.
	
	The equations of motion of the unicycle are summarized in Subsection~\ref{subsec:eqm} 
	and these are transformed to a path following reference frame in Subsection~\ref{subsec:pathFollowTrans} channeling to the control design of the next section.%

	\subsection{Equations of motion of the unicycle}\label{subsec:eqm}
	The first step of the Appellian approach is the same as the Lagrangian one, namely,
	${n_q = N-n_{\mathrm{g}} = 7}$ generalized coordinates ${q_i, i=1,\dots,n_q}$ have to be chosen that unambiguously represent the spatial configuration of the system.
	Let these be 
	\begin{equation}\label{eq:gen_coords}
		x_{\mathrm C}, \ \  y_{\mathrm C}, \ \  \psi, \ \  \vartheta, \ \  \varphi, \ \  \gamma, \ \  r.
	\end{equation}
	
	The second step is to define ${n_\sigma = N-n_{\mathrm g}-n_{\mathrm k} = 5}$ pseudovelocities ${\sigma_j, j=1,\dots,n_\sigma}$ such that the velocities and angular velocities can be unambiguously calculated.
	Here let the pseudovelocities be
	\begin{equation}\label{eq:pseudoVelocDefinition}
		\begin{split}
			\omega_{1} &:= \dot \vartheta,
			\\
			\omega_{2} &:= \dot \varphi +  \dot \psi \sin{\vartheta} ,
			\\
			\omega_{3} &:= \dot \psi \cos{\vartheta} ,
			\\
			\sigma_r &:= \dot r - \dot \vartheta R,
			\\
			\sigma_{\gamma} &:=    \dot \gamma h + \dot \varphi R \cos{\gamma} +  \dot \psi \big( h + R \cos{\gamma} \big) \sin{\vartheta} .
		\end{split}
	\end{equation}
	The pseudovelocities $\omega_{1},\omega_{2}$, and $\omega_{3}$ are the angular velocity components of the wheel resolved in frame~$\mathcal F_2$, 
	which worked well when deriving the equation of motion of the rolling wheel \citep{vizi2025tcst}.
	The pseudovelocity $\sigma_r$ is the $y_2$ component of the velocity $\mathbf v_\mathrm{A}$ of the point mass $m_1$; see \eqref{eq:vA_aA} in Appendix~\ref{app:govEqn}.
	The pseudovelocity  $\sigma_{\gamma}$ is the $x_3$ component of the velocity  $\mathbf v_\mathrm{B}$ of the point mass $m_2$; see \eqref{eq:vB_aB}.
	Such nontrivial choices significantly reduce the algebraic complexity of the resulting equations of motion.

	Taking into account the kinematic constraints \eqref{eq:kinemConstraints}, the derivatives of the general coordinates~\eqref{eq:gen_coords} can be expressed in terms of pseudovelocities~\eqref{eq:pseudoVelocDefinition} as
	\begin{equation} \label{eq:kinematicEQS}
		\begin{split}
			\dot x_{\mathrm C} &= \omega_{1}R  \sin{\psi} \cos{\vartheta} + \omega_{2}  R \cos{\psi}
			,\\
			\dot y_{\mathrm C} &= - \omega_{1} R  \cos{\psi} \cos{\vartheta} + \omega_{2}  R  \sin{\psi}
			,\\
			\dot \psi &= \frac{\omega_{3} }{\cos{\vartheta}}
			,\\
			\dot \vartheta &= \omega_{1}
			,\\
			\dot \varphi &= \omega_{2} - \omega_{3} \tan{\vartheta}
			,\\
			\dot r &= \sigma_r + \omega_{1} R  ,
			,\\
			\dot \gamma &={\sigma_{\gamma}}\frac{1}{h} - \omega_{2} \frac{R}{h}  \cos{\gamma} - \omega_{3} \tan{\vartheta},
		\end{split}
	\end{equation}
	assuming that the wheel is not horizontal, i.e., ${\vartheta\neq \pm\pi/2}$, which is excluded from the subsequent analysis.

	The further steps require one to derive the acceleration energy and the pseudoforces, which are detailed in Appendix~\ref{app:govEqn}.
	The Appellian approach yields
	\begin{equation} \label{eq:dynamicEQS}
			\mathbf M(\mathbf q)\,\dot{\bm \sigma} + \mathbf C(\mathbf q, \bm \sigma) = \bm \Pi(\mathbf q, \mathbf u)\,,
	\end{equation}
	where the vector 
	${\mathbf q = \big[x_{\mathrm C}\ \  y_{\mathrm C}\ \  \psi\ \  \vartheta\ \  \varphi\ \  \gamma \ \  r \big]^{\mathsf{T}}}$ collects the ge\-ne\-ralized coordinates, the vector
	${\bm \sigma = \big[\omega_{1}\ \ \omega_{2}\ \ \omega_{3}\ \ \sigma_r\ \ \sigma_{\gamma}\big]^{\mathsf{T}}}$ collects the pseudovelocities,
	${ \mathbf M(\mathbf q)}\in\mathbb R^{n_\sigma\times n_\sigma}$ is the generalized mass matrix (see \eqref{eq:massMtrx}, \eqref{eq:massMtrxElements}),
	while the vectors
	\begin{equation}\label{eq:CandPI}
		\begin{split}
			\mathbf C(\mathbf q, \bm \sigma) &= \big[C_{1}\ \ C_{2}\ \ C_{3}\ \ C_{4}\ \ C_{5}\big]^{\mathsf{T}}\,,
			\\
			\bm \Pi(\mathbf q, \mathbf u) &= \big[\Pi_{1}\ \ \Pi_{2}\ \ \Pi_{3}\ \ \Pi_{4}\ \ \Pi_{5}\big]^{\mathsf{T}}\,,
		\end{split}
	\end{equation}
	collect inertial forces (see \eqref{eq:vecInertialForcesElements}) 
	and the pseudoforces (see \eqref{eq:pseudo_forces}) where the latter contain the control inputs ${\mathbf u = [F \ \ T]^{\mathsf T}}$.
	One may notice that \eqref{eq:dynamicEQS} is quite similar to the common form of equations of motion used for holonomic robotic systems,
	however, it consists fewer equations, only ${n_\sigma=5}$ instead of ${n_q=7}$.
	From \eqref{eq:dynamicEQS}, the pseudo-accelerations can be expressed as
	\begin{equation}\label{eq:dsigmaDynEQS}
		\dot{\bm \sigma} = \mathbf M^{-1}(\mathbf q) \,\big( \bm \Pi(\mathbf q, \mathbf u) - \mathbf C(\mathbf q, \bm \sigma) \big)\,,
	\end{equation}
	since the mass matrix is nonsingular for positive masses.
	The dynamical equations~\eqref{eq:dsigmaDynEQS} together with the kinematic equations~\eqref{eq:kinematicEQS} form the equations of motion of the unicycle.

	The wheel-ground contact force
	\begin{equation} \label{eq:contactForce}
		\begin{split}
			\mathbf{K} &= %
            \begin{bmatrix}
            K_x \\ K_y \\ K_z
            \end{bmatrix}_{\mathcal F_1},
		\end{split}
	\end{equation}
	acts on the wheel, see Figure~\ref{fig:motivation}.
	To satisfy the corresponding kinematic and geometric constraints \eqref{eq:kinemConstraints} and \eqref{eq:geom_constr_eq_z}, this contact force must be within the friction cone, that is,
	\begin{equation}\label{eq:dynCondRoll}
		\mu_0 \geq \frac{\sqrt{K_x^2+K_y^2}}{K_z} \,, 
		\qquad K_z>0\,,
	\end{equation}
	where $\mu_0$ is the static friction coefficient.
	While the Appell equations do not provide information about the contact force, this can be obtained using free body diagrams and applying the Newton--Euler approach, as shown in Appendix~\ref{app:NewtonEuler_contactForce}.
	These yield~\eqref{eq:contactForceComps}.
	Then, utilizing the solution of~\mbox{\eqref{eq:kinematicEQS}--\eqref{eq:dsigmaDynEQS}}, one may compute the required static friction coefficient needed to maintain the rolling without slipping.

	\subsection{Path-following representation}\label{subsec:pathFollowTrans}
	
\begin{figure}[!t]
		\centering
		\includegraphics[width=0.75\columnwidth]{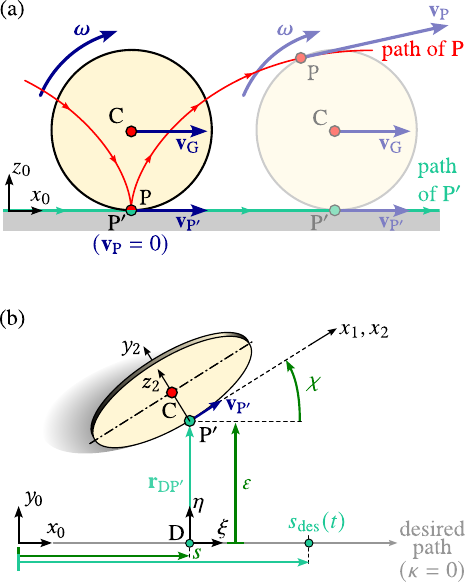}
		\caption{(a) Illustration of the velocity pole $\mathrm P^\prime$ and pole changing velocity $\mathbf v_{\mathrm P^\prime}$, (b) path-following coordinate transformation}
		\label{fig:pathFollowCrdTransform}
	\end{figure}

	When discussing the mechanics of rolling, the material point $\mathrm P$ of the wheel contacting the ground and the geometric wheel-ground contact point $\mathrm P^\prime$ must be distinguished. 
	As the point $\mathrm P$ is a material point on the circumference of the wheel, it touches the ground at a given time instance, 
	at which time it coincides with the geometric wheel-ground contact point $\mathrm P^\prime$, see Figure~\ref{fig:pathFollowCrdTransform}(a).
	At this time instance the kinematic condition of rolling without slipping is formulated for the material point $\mathrm P$ as  ${\mathbf{v}_{\mathrm P} = \mathbf 0}$, see \eqref{eq:vP}. 
	In the next time instance, as the wheel rolls forward, $\mathrm P$ lifts up from the ground and the neighboring material point on the circumference of the wheel becomes the contact point $\mathrm P^\prime$.
	As time goes, the contact point $\mathrm P^\prime$ moves forward and its path can be imagined as the trail of the rolling wheel on the ground with nonzero velocity, i.e., ${\mathbf v_{\mathrm P^\prime}\neq \mathbf 0}$.
	The circumferential points of the wheel form the so-called moving polode, while the trail of the wheel on the ground is called the fixed polode. An interested reader can find more details on this topic in \citep{csernak2019pole}.
	
	The position of the contact point $\mathrm P^\prime$ is given by
	\begin{equation}\label{eq:geomPolePos}	
		\mathbf r_{{\mathrm P'}}
			= \mathbf r_{\mathrm C} + \mathbf r_{\mathrm C {\mathrm P'}} ,
	\end{equation}
	that yields
	\begin{equation}\label{eq:geomPolePos2}
		\begin{bmatrix}
			x_{{\mathrm P'}}
			\\ y_{{\mathrm P'}}
			\\ 0
		\end{bmatrix}_{{\mathcal F}_{0}} 
		=
		\begin{bmatrix}
			x_{\mathrm{C}} \\ 
			y_{\mathrm{C}} \\
			z_{\mathrm{C}}
		\end{bmatrix}_{{\mathcal F}_{0}} 
		\!\!+
		\left[\begin{matrix} 
			- R \sin{\psi} \sin{\vartheta} \\ 
			\ R \cos{\psi} \sin{\vartheta}  \\ 
			- R \cos{\vartheta} 
		\end{matrix}\right]_{{\mathcal F}_{0}}
		\!\!,
	\end{equation}
	where we used \eqref{eq:positionG} and transformed ${\mathbf r_{\mathrm{C}{\mathrm P'}} = \big[0\ \ 0\ \ -R\big]_{\mathcal{F}_2}^\top}$ to the $\mathcal{F}_0$ frame.
	Taking the time derivative, the pole changing velocity can be calculated as
	\begin{equation}\setlength\arraycolsep{1ex}
		\mathbf v_{{\mathrm P'}} 
		= \dot{\mathbf r}_{{\mathrm P'}}
		= \dot{\mathbf r}_{\mathrm{C}}  + \dot{\mathbf r}_{\mathrm{C}{\mathrm P'}} ,
	\end{equation}
	yielding 
	\begin{equation}\label{eq:vpCalcStep}
		\begin{bmatrix}
			\dot x_{{\mathrm P'}} \\ 
			\dot y_{{\mathrm P'}} \\ 
			0
		\end{bmatrix}_{{\mathcal F}_{0}} 
		=
		\begin{bmatrix}
			\dot x_{\mathrm{C}} \\ 
			\dot y_{\mathrm{C}}	\\ 
			\dot z_{\mathrm{C}}
		\end{bmatrix}_{{\mathcal F}_{0}} 
		\!\!+
		\left[\begin{matrix} 
			- \dot \psi R  \cos{\psi} \sin{\vartheta} -  \dot \vartheta R \sin{\psi} \cos{\vartheta} \\ 
			- \dot \psi  R \sin{\psi} \sin{\vartheta} + \dot \vartheta R \cos{\psi} \cos{\vartheta}  \\ 
			\dot \vartheta R \sin{\vartheta} 
		\end{matrix}\right]_{{\mathcal F}_{0}}
		\!\!. 
	\end{equation}
	Substituting ${\dot x_{\mathrm{C}}, \dot y_{\mathrm{C}}}$ from the kinematic constraints \eqref{eq:kinemConstraints} and $\dot \psi, \dot\vartheta,\dot\varphi$ from the pseudovelocities \eqref{eq:kinematicEQS} gives
	\begin{equation}\label{eq:vPprime}
		\begin{split}
			\dot x_{{\mathrm P'}} &= \big( \omega_{2} - \omega_{3}  \tan{\vartheta} \big)\, R \cos{\psi} ,
			\\
			\dot y_{{\mathrm P'}} &= \big( \omega_{2} - \omega_{3}  \tan{\vartheta} \big)\, R \sin{\psi} .
		\end{split}
	\end{equation}
	
	With this coordinate transformation, the wheel position can be characterized by the wheel-ground contact ${\mathrm P'}$ instead of the wheel center $\mathrm C$. 
	That is, in \eqref{eq:gen_coords} the coordinates ${x_{\mathrm C}, y_{\mathrm C}}$ are substituted with ${x_{{\mathrm P'}}, y_{{\mathrm P'}}}$ yielding the new set of generalized coordinates 
	\begin{equation}\label{eq:gen_coords_with_geomPole}
		x_{{\mathrm P'}}, \ \  y_{{\mathrm P'}}, \ \  \psi, \ \  \vartheta, \ \  \varphi, \ \  \gamma, \ \  r.
	\end{equation}
	Consequently, ${\dot x_{{\mathrm P'}}, \dot y_{{\mathrm P'}}}$ given in \eqref{eq:vPprime} replace the expressions of  ${\dot x_{\mathrm C}, \dot y_{\mathrm C}}$ in the equations of motion~\eqref{eq:kinematicEQS} while the other five equations remain the same as these come from the definitions of pseudovelocities \eqref{eq:pseudoVelocDefinition}.
	
	We assume that a desired path is defined as a function of the arc length $s$ and, 
	following \citep{qin2022nonholonomic}, we transform
	the system to a path-reference frame.
	Namely, along the desired path we consider a geometric point $\mathrm D$ whose position is given by
	\begin{equation}
		\mathbf r_{\mathrm D} = 
        \begin{bmatrix}
        x_{\mathrm D}(s) \\  y_{\mathrm D}(s) \\  0
        \end{bmatrix}_{\mathcal{F}_0},
	\end{equation}
    see Figure~\ref{fig:pathFollowCrdTransform}(b).
	The path reference frame $\mathcal{F}_\mathrm{p}$ is laid on the ground with coordinate axes $\xi,\eta,\zeta$ and origin $\mathrm D$ such that $\xi$ is tangential to the path and $\eta$ points toward $\mathrm P^\prime$.
	The lateral error, that is, the signed distance of points $\mathrm D$ and $\mathrm P^\prime$ is denoted by $\varepsilon$, 
	which yields
	\begin{equation}
		\mathbf r_{\mathrm D{\mathrm P'}} = 
        \begin{bmatrix}
        0 \\ \varepsilon \\ 0
        \end{bmatrix}_{\mathcal F_\mathrm{p}}.
	\end{equation}
	The desired yaw angle, that is, the angle of the $x_0$ and $\xi$ axes are denoted by $ \psi_{\mathrm D}(s)$.
	Accordingly, the yaw angle error is defined as 
	\begin{equation}\label{eq:chi}
		\chi = \psi - \psi_{\mathrm D}(s)\,.
	\end{equation}
	Finally, the signed curvature of the desired path at point D is given by
	\begin{equation}\label{eq:kappa}
		\kappa(s) =  \frac{\mathrm d \psi_{\mathrm D}}{\mathrm d s} .
	\end{equation}

	The absolute position of ${\mathrm P'}$ can be obtained as
	\begin{equation}
		\mathbf r_{{\mathrm P'}} = 	
		\mathbf r_{\mathrm D} + \mathbf r_{\mathrm D{\mathrm P'}} ,
	\end{equation}
	which yields
	\begin{equation}
			\begin{bmatrix}
				x_{\mathrm P^{\prime}}\\
				y_{\mathrm P^{\prime}}\\
				0
			\end{bmatrix}_{{\mathcal F}_{0}} 
			= \begin{bmatrix}
				x_{\mathrm D}{\left(s \right)} - \varepsilon \sin{\psi_{\mathrm D}{\left(s \right)} } 
				\\
				y_{\mathrm D}{\left(s \right)} + \varepsilon \cos{\psi_{\mathrm D}{\left(s \right)} } 
				\\
				0
			\end{bmatrix}_{{\mathcal F}_{0}} .
	\end{equation}
	Taking the derivative the velocity of ${\mathrm P'}$ becomes 
	\begin{equation}
		\mathbf v_{{\mathrm P'}} =\dot{\mathbf r}_{{\mathrm P'}}  = 
		\dot{\mathbf r}_{\mathrm D} + \dot{\mathbf r}_{\mathrm D{\mathrm P'}} ,
	\end{equation}
	which yields
	\begin{equation}\label{eq:conn_sec_xyp}
		\begin{bmatrix}
			\dot x_{\mathrm P^{\prime}}
			\\
			\displaystyle \dot y_{\mathrm P^{\prime}}
			\\
			0
		\end{bmatrix}_{{\mathcal F}_{0}} \!\!\!\! = \begin{bmatrix}
			\dot s \left(1- \kappa{\left(s \right)} \varepsilon \right) \cos{\psi_{\mathrm D}{\left(s \right)} }  - \dot \varepsilon \sin{\psi_{\mathrm D}{\left(s \right)} } 
			\\
			\dot s \left(1 - \kappa{\left(s \right)} \varepsilon \right) \sin{\psi_{\mathrm D}{\left(s \right)} }  + \dot \varepsilon \cos{\psi_{\mathrm D}{\left(s \right)} } 
			\\
			0
		\end{bmatrix}_{{\mathcal F}_{0}},
	\end{equation}
	where the identities 
	\begin{equation}
		\begin{split}
			\dot x_{\mathrm D} &=  \frac{\mathrm d x_{\mathrm D}}{\mathrm d t }  = \frac{\mathrm d x_{\mathrm D}}{\mathrm d s} \frac{\mathrm d s}{\mathrm d t}  = \dot s \cos{\psi_{\mathrm D}(s) }\,, 
			\\
			\dot y_{\mathrm D} &=  \frac{\mathrm d y_{\mathrm D}}{\mathrm d t }  = \frac{\mathrm d y_{\mathrm D}}{\mathrm d s} \frac{\mathrm d s}{\mathrm d t} = \dot s \sin{\psi_{\mathrm D}(s) }\,, 
			\\
			\dot \psi_{\mathrm D} &=  \frac{\mathrm d \psi_{\mathrm D}}{\mathrm d t }  = \frac{\mathrm d \psi_{\mathrm D}}{\mathrm d s} \frac{\mathrm d s}{\mathrm d t} = \dot s \,\kappa(s)\,,
		\end{split}
	\end{equation}
	are applied, see \eqref{eq:kappa}. 
	Also, taking the derivative of \eqref{eq:chi} yields
	\begin{equation}\label{eq:chidot}
		\dot \chi = \dot \psi - \dot s\, \kappa{\left(s \right)}\,,
	\end{equation}
	and combining this with \eqref{eq:conn_sec_xyp}, the rates of the arc length, lateral error and yaw angle error are expressed as 
	\begin{equation}\label{eq:crdTransfGeneral}
		\begin{split}
			\dot s &= \frac{1 }{1- \kappa \varepsilon }\big(
			\dot x_{{\mathrm P'}} \cos{{\psi_{\mathrm{D}} }}  + \dot y_{{\mathrm P'}} \sin{{\psi_{\mathrm{D}} }} 
			\big)\,,
			\\
			\dot \varepsilon &= - \dot x_{{\mathrm P'}}  \sin{{\psi_{\mathrm{D}} }} + \dot y_{{\mathrm P'}} \cos{{\psi_{\mathrm{D}} }}\,,
			\\
			\dot \chi &= \dot \psi - \frac{ \kappa}{1- \kappa \varepsilon}\big( \dot x_{{\mathrm P'}} \cos{{\psi_{\mathrm{D}} }}  + \dot y_{{\mathrm P'}} \sin{{\psi_{\mathrm{D}} }} \big)\,.
		\end{split}
	\end{equation}%

	Substituting ${\dot x_{{\mathrm P'}}, \dot y_{{\mathrm P'}}}$ from \eqref{eq:vPprime} and using ${\dot \vartheta, \dot \varphi, \dot r, \dot \gamma}$ from \eqref{eq:kinematicEQS} yield the kinematic equations 
	\begin{equation} \label{eq:kinematicEQS_withPATH}
		\begin{split}
			\dot s &= \frac{1 }{1- \kappa \varepsilon } \big(
			\omega_{2}  - \omega_{3}   \tan{\vartheta}
			\big)\, R  \cos{\chi },
			\\
			\dot \varepsilon &= \big( \omega_{2}  -  \omega_{3}  \tan{\vartheta} \big)\, R \sin{\chi } ,
			\\
			\dot \chi &= \frac{\omega_{3} }{ \cos{\vartheta} } - \frac{\kappa }{ 1- \kappa \varepsilon } \big(
			\omega_{2} - \omega_{3}  \tan{\vartheta} %
			\big) \, R \cos{\chi } ,
			\\
			\dot \vartheta &= \omega_{1} ,
			\\
			\dot \varphi &= \omega_{2} - \omega_{3} \tan{\vartheta}\,,
			\\
			\dot r &= \sigma_r + \omega_{1} R\,,
			\\
			\dot \gamma &={\sigma_{\gamma}}\frac{1}{h} - \omega_{2} \frac{R}{h}  \cos{\gamma} - \omega_{3} \tan{\vartheta}\,,
		\end{split}
	\end{equation}
	corresponding to the set of generalized coordinates 
	\begin{equation}\label{eq:gen_coords_with_PATH}
		s, \ \ \varepsilon, \ \ \chi, \ \ \vartheta, \ \  \varphi, \ \  \gamma, \ \  r.
	\end{equation}
	The first two equations in \eqref{eq:kinematicEQS_withPATH} give $\dot{s}$ and $\dot\varepsilon$ and these represent the kinematic conditions of rolling  formalized with the new path coordinates.
	Meanwhile, the last five equations in \eqref{eq:kinematicEQS_withPATH}, 
	enable one to express the pseudovelocities $\omega_1$, $\omega_2$, $\omega_3$, $\sigma_r$, $\sigma_\gamma$ using the generalized velocities $\dot\chi$, $\dot\theta$, $\dot\varphi$, $\dot{r}$, $\dot\gamma$.

	The equations of motion of the unicycle in the path-following coordinates consist of \eqref{eq:dsigmaDynEQS} and \eqref{eq:kinematicEQS_withPATH}. 
	These comply the control affine form 
	\begin{equation}\label{eq:1stOrderEQS}
		\dot{\mathbf x} = f(\mathbf x) + g(\mathbf x) \mathbf u ,
	\end{equation}
	with state and input vectors
	\begin{equation}
		\begin{split}
			\mathbf x &= \big[
			\omega_{1} 
			\ \ \ \sigma_r 
			\ \ \  r 
			\ \ \ \vartheta 
			\ \ \ \omega_{3} 
			\ \ \ \chi 
			\ \ \ \varepsilon 
			\ \ \ \omega_{2} 
			\ \ \ \sigma_{\gamma} 
			\ \ \ \gamma 
			\ \ \ \varphi 
			\ \ \  s\big]^{\mathsf T}\,,
			\\
			\mathbf u &= \big[F \ \ T\big]^{\mathsf{T}}\,,
		\end{split}
	\end{equation}
	respectively, where the states are ordered to simplify the control design presented in the next section.

\section{Motion Planning and Control Design}\label{sec:controlDesign}

Motion planning and control for unicycles requires special attention due to their strongly nonlinear and highly unstable dynamical characteristics.
First we perform path-planning in Subsection~\ref{sec:Maneuver} followed by the control design in Subsection~\ref{sec:control}.
These will be integrated later in Section~\ref{sec:simulation}.

\subsection{Maneuver planning using clothoids}\label{sec:Maneuver}

As mentioned above, we distinguish the kinematic unicycle model and the dynamic models of real physical unicycles. 
The former one is widely used for path planning of wheeled robots~\citep{Bianco2004Smooth,Giordano2009Shortest,Deptula2020Approximate,Ben2022time,Consolini2009Stabilization,Morro2011Path,Becker2012Approximate,Salaris2015Epsilon,Li2022source,dubins1957curves,qin2022nonholonomic}.
Here we use this model to design the desired path of the point $\mathrm D$ that we want the geometric wheel-ground contact point $\mathrm P^\prime$ to follow, see Figure~\ref{fig:pathFollowCrdTransform}(b).
    
The kinematic unicycle model can be written as
\begin{equation}\label{eq:kinemunic}
	\begin{split}
		\dot x &= v\cos\psi,
        \\
		\dot y &= v\sin\psi,
        \\
		\dot \psi &= \omega,
	\end{split}
\end{equation}
where $x$ and $y$ denote the position of the contact point in the ground plane, $\psi$ gives the heading direction, $v$ is the speed, $\omega$ is the angular velocity,
and the dot denotes the derivatives respect to time $t$.
Considering arc-length parametrization~\citep{oh2025sharable} the model can be rewritten as
\begin{equation}\label{eq:kinemunic_arclength_diffeq}
	\begin{split}
		x'    &= \cos\psi,
        \\
		y'    &= \sin\psi,
        \\
		\psi' &= \kappa,
	\end{split}
\end{equation}
where the prime denotes the derivatives respect to the arc length $s$, while the curvature $\kappa(s)$ encodes the path as a function of the arc length.

We divide the path into straight and turning sections such that acceleration/deceleration are only planned during the straight sections.

The geometric design of straight sections is trivial as the desired curvature is given by
\begin{equation}
	\kappa_{\mathrm{des}}(s)\equiv0\,, \quad s\in\big[s_{\mathrm i}\,, \,s_{\mathrm{f}}\big],
\end{equation} 
where ${s_{\mathrm i} = s(t_{\mathrm i})}$ 
and ${s_{\mathrm{f}} = s(t_{\mathrm{f}})}$ are the initial and final arc lengths at the beginning and at the end of the straight section, respectively. 
The boundary conditions for a straight section are
\begin{equation}\label{eq:bondaryCondsStraight}\arraycolsep=1pt
	\begin{array}{rlrlrlrl}
		x(s_{\mathrm i})         &= x_{\mathrm i},           ~~~
		&y(s_{\mathrm i})        &= y_{\mathrm i},           ~~~
		&\psi(s_{\mathrm i})     &= \psi_{\mathrm i},        ~~~
		&\kappa(s_{\mathrm i})   &=  0 ,
		\\
		x(s_{\mathrm{f}} )        &= x_{\mathrm{f}},	 ~~~
		&y(s_{\mathrm{f}}  )      &= y_{\mathrm{f}},	 ~~~
		&\psi(s_{\mathrm{f}})     &= \psi_{\mathrm{f}},  ~~~
		&\kappa(s_{\mathrm{f}})   &=  0,
	\end{array}
\end{equation}
	with ${\psi_{\mathrm i} = \psi_{\mathrm{f}} = \mathrm{atan}_2\big(y_{\mathrm{f}}-y_{\mathrm i} , x_{\mathrm{f}}-x_{\mathrm i}\big)}$.
	Using the two-ar\-gu\-ment arctangent function instead of the regular single-ar\-gu\-ment one makes correct quadrant tracking and makes it possible to move parallel to the $y$ axis.
	The length of the straight section is
\begin{equation}
		\Delta s = s_{\mathrm{f}}-s_{\mathrm i} = \sqrt{ (x_{\mathrm{f}}-x_{\mathrm i})^2 + (y_{\mathrm{f}}-y_{\mathrm i})^2}.
\end{equation}

The smooth velocity profile 
\begin{equation}\label{eq:vdesdesign}
	v_{\mathrm{des}}(t) = v_{\mathrm i} + \dfrac{\Delta v}{2} \bigg( 1- \cos\frac{\pi(t -  t_{\mathrm i})}{{\Delta t}} \bigg) \,, \quad t\in\big[t_{\mathrm i}\,, \,t_{\mathrm{f}}\big].
\end{equation}
is assumed where ${ v_{\mathrm i} = v_{\mathrm{des}}(t_{\mathrm i}) }$ 
and  ${ v_{\mathrm{f}} = v_{\mathrm{des}}(t_{\mathrm{f}}) }$  are the initial and final velocities, 
${\Delta v = v_{\mathrm{f}}-v_{\mathrm i}}$ is the desired velocity change, 
while ${
	\Delta t =t_{\mathrm{f}} - t_{\mathrm i} 
}$ is the desired time duration along the straight section.
Integrating the velocity profile yields the desired arc length profile
	\begin{equation}
		s_{\mathrm{des}}(t) =  s_{\mathrm i}+\left(v_{\mathrm i}\!+\!\frac{{\Delta{}v} }{2}\right) \! \left(t \! -\!  t_{\mathrm i}\right) - \frac{ {\Delta{}v}\,\Delta t }{2 \pi}\sin{\frac{\pi \left(t \! - \! t_{\mathrm i}\right)}{ \Delta t} }, \label{eq:sdesdesign}
	\end{equation}
where ${s_{\mathrm{des}}(t_{\mathrm i})=s_{\mathrm i}}$
and ${s_{\mathrm{des}}(t_{\mathrm{f}})=s_{\mathrm{f}}}$.

The turning sections are designed using three-clothoid segments~\citep{bertolazzi2018g2,oh2025sharable}. 
The prominent advantages of this approach are as follows: 
(i) it has a small number of parameters, which all have clear physical meanings; 
(ii) it has 2 tunable parameters, which allow the generation of multiple paths for given boundary conditions;
(iii) the resulting paths are $G^2$ continuous; 
(iv) the numerical solution exactly satisfies the 8 boundary conditions. 

The path curvature is given as
\begin{equation}\label{eq:kappaDesTurn}\arraycolsep=1pt
	\kappa_{\mathrm{des}}(s) = 
	\begin{cases}
		\begin{array}{llrl}
			\kappa_{\mathrm{des},1}(s), & ~~~\text{if  } s_{\mathrm i}   \leq s \leq s_{\mathrm i} + s_0,\\
			\kappa_{\mathrm{des},2}(s), & ~~~\text{if  } s_{\mathrm i} + s_0 \leq s \leq s_{\mathrm i} + s_0+s_1,\\
			\kappa_{\mathrm{des},3}(s), & ~~~\text{if  } s_{\mathrm i} + s_0+s_1 \leq s \leq s_{\mathrm f},
		\end{array}
	\end{cases}
\end{equation}
where 

\begin{equation}\label{eq:kappaDesTurn2}
	\begin{split}
		\kappa_{\mathrm{des},1}(s) &= \kappa_{\mathrm i} + \kappa'_0 (s - s_{\mathrm i}),
		\\
		\kappa_{\mathrm{des},2}(s) &= \kappa_\mathrm{m} + \kappa'_1 (s- s_{\mathrm i}-s_0-s_1/2),
		\\
		\kappa_{\mathrm{des},3}(s) &= \kappa_{\mathrm{f}} + \kappa'_2 (s-s_{\mathrm{f}}),
	\end{split}
\end{equation}
and ${s_{\mathrm f} = s_{\mathrm i} + s_0+s_1+s_2}$ is the arc length at the end of the turning.
The sign of $\kappa(s)$ gives the direction of the turn, ${\kappa>0}$ and ${\kappa<0}$ mean a left and a right turn, respectively.

The corresponding positions and heading angle at arc length $s$ are
\vspace{-1ex} 
	\begin{equation}\label{eq:headingAndPosXY_clothTurn}
	\begin{split}
		\psi(s) &= \psi_{\mathrm i} + \int_{s_{\mathrm i}}^{s} \kappa(\sigma)\,\mathrm d \sigma,
		\\
		x(s) &= x_{\mathrm i} + \int_{s_{\mathrm i}}^{s} \cos\psi(\sigma)\, \mathrm d\sigma,
		\\
		y(s) &= y_{\mathrm i} + \int_{s_{\mathrm i}}^{s}  \sin\psi(\sigma)\, \mathrm d\sigma, 
	\end{split}	 
\end{equation}
cf.~\eqref{eq:kinemunic_arclength_diffeq}. 
This parametrization has 18 parameters in total, which consists of 8 known boundary conditions
\begin{equation}\label{eq:bondaryConds}\arraycolsep=1pt
	\begin{array}{rlrlrlrl}
		x(s_{\mathrm i})         &= x_{\mathrm i},           ~~~
		&y(s_{\mathrm i})        &= y_{\mathrm i},           ~~~
		&\psi(s_{\mathrm i})     &= \psi_{\mathrm i},        ~~~
		&\kappa(s_{\mathrm i})   &= \kappa_{\mathrm i} ,
		\\
		x(s_{\mathrm{f}} )        &= x_{\mathrm{f}},	       ~~~
		&y(s_{\mathrm{f}}  )      &= y_{\mathrm{f}},	       ~~~
		&\psi(s_{\mathrm{f}})     &= \psi_{\mathrm{f}},        ~~~
		&\kappa(s_{\mathrm{f}})   &= \kappa_{\mathrm{f}},
	\end{array}
\end{equation}
and 10 unknown parameters
\begin{equation}\label{eq:cloth_unkwn}
	s_0,\ \,\kappa'_0,\ \,s_1,\ \,x_\mathrm{m},\ \,y_\mathrm{m},\ \,\psi_\mathrm{m},\ \,\kappa_\mathrm{m},\ \,\kappa'_1,\ \,s_2,\ \,\kappa'_2,
\end{equation}
where the subscript $\mathrm m$ indicated the variables ${x,y,\psi,\kappa}$ at 
\begin{equation}
	{s_\mathrm{m}=s_{\mathrm i}+s_0+s_1/2},
\end{equation}
that is, at the midpoint of the second clothoid segment.

The curvature $\kappa(s)$ in \eqref{eq:kappaDesTurn}, the heading angle $\psi(s)$ and positions  $x(s)$ and $y(s)$ in \eqref{eq:headingAndPosXY_clothTurn} need to be continuous at the clothoid segment boundaries ${s=s_{\mathrm i} + s_0}$ and ${s=s_{\mathrm i} + s_0+s_1}$. 
These continuity equations result in 8 algebraic equations, see \citep{oh2025sharable}. 
{\par\nobreak\vfil\penalty0\vfilneg\vtop\bgroup
Using the continuity equations, 6 of the 10 unknowns in \eqref{eq:cloth_unkwn} can be algebraically eliminated, yielding
\begin{equation}
	\begin{split}
		\kappa^{\prime}_{0} &= \frac{2 \Delta\psi -\kappa^{\prime}_{1} \big( s_{1}^{2} + s_{1} s_{2}\big) - \kappa_{\mathrm{f}} s_{2} - \kappa_{\mathrm i} \big( 2 s_{0} + 2 s_{1} + s_{2}\big)}{s_{0} \left(s_{0} + 2 s_{1} + s_{2}\right)},
		\\
		\kappa^{\prime}_{2} &= \frac{\kappa_{\mathrm i} s_{0} - \kappa^{\prime}_{1} \big( s_{0} s_{1} + s_{1}^{2}\big) + \kappa_{\mathrm{f}} \big(s_{0} + 2 s_{1} + 2 s_{2}\big) - 2 \Delta\psi }{s_{2} \big(s_{0} + 2 s_{1} + s_{2}\big)},
		\\
		\kappa_\mathrm{m} &= \frac{ 4 \Delta\psi + \kappa^{\prime}_{1} \left(s_{0} s_{1} - s_{1} s_{2}\right) - 2 \kappa_{\mathrm{f}} s_{2} - 2 \kappa_{\mathrm i} s_{0}}{2 \left(s_{0} + 2 s_{1} + s_{2}\right)},
		\\
		\psi_\mathrm{m} &= \frac{1}{8 \big(s_{0} + 2 s_{1} + s_{2}\big)}\Big(
		8 \psi_{\mathrm{f}}( s_{0} + s_{1}) + 8 \psi_{\mathrm i}( s_{1} + s_{2})
		\\&\quad
		+ \kappa_{\mathrm i} \big(4 s_{0} s_{1} + 4 s_{0} s_{2}\big) 
		- \kappa_{\mathrm{f}} \big( 4 s_{0} s_{2} + 4 s_{1} s_{2}\big) 
		\\&\quad
		-
		\kappa^{\prime}_{1} \big(  3 s_{0} s_{1}^{2} + 4 s_{0} s_{1} s_{2} + 2 s_{1}^{3} + 3 s_{1}^{2} s_{2}\big) 
		\Big),
		\\
		x_\mathrm{m} &= x_{\mathrm i} + s_{0}           \mathrm C{\left(a_0,	b_0,	c_0 \right)}
		+ \frac{s_{1} }{2}\mathrm C{\left(a_1,	b_1,	c_1 \right)},
		\\
		y_\mathrm{m} &= y_{\mathrm i} + s_{0}           \mathrm S{\left(a_0,	b_0,	c_0 \right)}
		+ \frac{s_{1} }{2}\mathrm S{\left(a_1,	b_1,	c_1 \right)},
	\end{split}
\end{equation}
\par\xdef\tpd{\the\prevdepth}\egroup\prevdepth=\tpd}
\newpage
\noindent
where ${\Delta\psi = \psi_{\mathrm f} - \psi_{\mathrm i}}$ and we evaluate the Fresnel integrals
\begin{equation}
	\begin{split}
		\mathrm C(a,b,c) &= \int_{0}^{1}\cos\left(\dfrac{a}{2}\sigma^2 + b\sigma +c\right)\,\mathrm d \sigma\,,
		\\
		\mathrm S(a,b,c) &= \int_{0}^{1}\sin\left(\dfrac{a}{2}\sigma^2 + b\sigma +c\right)\,\mathrm d \sigma\,,
	\end{split}
\end{equation}
at 
\begin{equation}\label{eq:clothoidCfs}
	\arraycolsep=1pt
	\begin{array}{rlcrlcrl}
		a_0 & = \kappa^\prime_0 s_0^2, & \quad~ & b_0 & = \kappa_{\mathrm i} s_{0}, & \quad~ & c_0 & = \psi_{\mathrm i},   \\[1ex]
		a_1 & = \kappa^\prime_1 s_1^2/4, & \quad~ & b_1 & = \kappa_\mathrm{m} s_{1}/2, & \quad~ & c_1 & = \psi_\mathrm{m},        \\[1ex]
		a_2 & = \kappa^\prime_2 s_2^2, & \quad~ & b_2 & = -\kappa_{\mathrm{f}} s_{2}, & \quad~ & c_2 & = \psi_{\mathrm{f}}.
	\end{array}
\end{equation}
This way only the 2 nonlinear equations 
\begin{equation}\label{eq:clothoidFinalEqs}
	\begin{split}
		& s_{0}           \mathrm C{\left(a_0,	b_0,	c_0 \right)}
		+ \frac{s_{1} }{2}\mathrm C{\left(a_1,	b_1,	c_1 \right)} 
		+ \frac{s_{1} }{2}\mathrm C{\left(a_1,	-b_1,	c_1 \right)} 
		\\&\quad
		+ s_{2}           \mathrm C{\left(a_2,	b_2,	c_2 \right)} = x_{\mathrm{f}} - x_{\mathrm i}\,,
		\\
		& s_{0} \mathrm S{\left(a_0,		b_0,		c_0 \right)}
		+ \frac{s_{1} }{2}\mathrm S{\left(a_1,	b_1,	c_1 \right)}
		+ \frac{s_{1} }{2}\mathrm S{\left(a_1,	-b_1,	c_1 \right)}
		\\&\quad
		+ s_{2} \mathrm S{\left(a_2,	b_2,	c_2 \right)} = y_{\mathrm{f}} - y_{\mathrm i},
	\end{split}
\end{equation}
remain for the 4 unknowns ${s_0,s_1,s_2}$ and ${\kappa'_1}$.

For simplicity, we constraint the solution by setting ${s_0 = s_2 = \mathcal{S}s_1 }$, in which ${\mathcal S>0}$ is the tuning parameter that we call the dimensionless clothoid segment ratio. 
The special case ${\mathcal S=1}$ means that all clothoid segments have equal lengths.
Then \eqref{eq:clothoidFinalEqs} can be solved numerically for the remaining unknowns ${s_1}$ and ${\kappa^\prime_1}$.
Note that the solution may not be unique and numerical algorithms like
Newton's method may converge to one of these depending on the initialization.
In order to find all solutions, one may utilize the multidimensional bisection method \citep{bachrathy2012bisection}.

	As mentioned above, intentional velocity changes during the turning sections are not considered in this approach,
	so the longitudinal controller is ``switched to balancing mode'' by setting the desired velocity as the instantaneous velocity, i.e., assuming zero velocity error.
Nevertheless, the velocity may slightly change during the turning phase due to the strongly nonlinear dynamical nature of the unicycle.

\begin{figure*}[!t]
	\centering
	\includegraphics[width=0.666\textwidth]{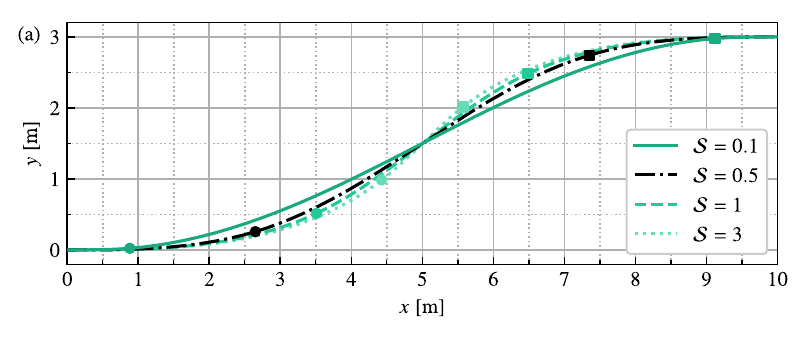}%
	\includegraphics[width=0.333\textwidth]{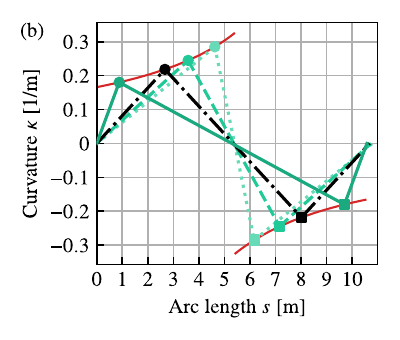}
	\caption{%
		(a) Lane change with ${\Delta x = 10}$\,m, ${\Delta y=3}$\,m, and ${\Delta \psi=0^\circ}$ for various clothoid section ratios
		${\mathcal{S} =  s_0/s_1 = s_2/s_1 }$. %
		Dots mark the boundaries of the first and second clothoid segments, while squares mark the boundaries of the second and third segments.
		(b) The curvature along the lane change.
		The thin red curves show the extrema of the curvatures in case of varying $\mathcal S$.%
	}%
	\label{fig:clothSegmRatioEffect}
\end{figure*}

The path planning algorithm is demonstrated on a lane change maneuver.
Initially the unicycle is at standstill, that is, ${v_\mathrm{i}=0}$, at the origin with heading along the $x_0$~axis, that is, ${\psi_\mathrm{i} = 0}$.
The maneuver consists of three parts.

The first part is a straight segment during which the unicycle aims to accelerate from standstill to ${v_\mathrm{f}=1.5}$\,m/s within the distance ${\Delta s=5}$\,m  along the $x$~axis, i.e., ${\kappa_{\mathrm{des}}(s) \equiv 0}$.
	This takes ${\Delta{}t = \tfrac{20}{3}\,\mathrm{s}\approx 6.666}$\,s with the desired velocity  \eqref{eq:vdesdesign} and arc length \eqref{eq:sdesdesign} profiles becoming
\begin{equation}
	\begin{split}
		v_{\mathrm{des}}(t) &= 0.75- 0.75\cos\big({0.15 \pi t}\big) \,,
		\\
		s_{\mathrm{des}}(t) &=  0.75 t - \frac{5}{\pi}\sin\big({0.15 \pi t}\big)\,, \quad t\in\big[0\,, \,6.666\big]\,\mathrm{s}.
	\end{split}
\end{equation}

The second part is a lane change with length ${\Delta x=10}$\,m,
width ${\Delta y=3}$\,m and orientation change ${\Delta \psi=0}$.
Considering the straight part before the lane change, the corresponding boundary conditions \eqref{eq:bondaryConds} become
\begin{equation}\arraycolsep=1pt
	\begin{array}{rlrlrlrl}
		x(s_{\mathrm i})         &= 5\,\mathrm{m},      ~~~
		&y(s_{\mathrm i})        &= 0\,\mathrm{m},      ~~~
		&\psi(s_{\mathrm i})     &= 0,            ~~~
		&\kappa(s_{\mathrm i})   &= 0 ,
		\\
		x(s_{\mathrm{f}} )        &= 15\,\mathrm{m},	 ~~~
		&y(s_{\mathrm{f}}  )      &= 3\,\mathrm{m},	     ~~~
		&\psi(s_{\mathrm{f}})     &= 0,            ~~~
		&\kappa(s_{\mathrm{f}})   &= 0,
	\end{array}
\end{equation}
where the starting arc length is $s_{\mathrm i}=5$\,m while the arc length $ s_{\mathrm{f}} $ at the end of the turning part is unknown a priori.
The family of lane changes is plotted with clothoid segment ratios $\mathcal S = 0.1, 0.5, 1, 3$ in Figure~\ref{fig:clothSegmRatioEffect}(a) while panel (b) presents the corresponding curvature profiles.

The clothoid segment ratio ${\mathcal S=0.5}$ yield a special lane change for which the sharpnesses for the three clothoid segments are of the same magnitude, namely, ${\kappa'_{0} \equiv-\kappa'_{1}\equiv\kappa'_{2}}$.
Using ${\mathcal S=0.5}$ gives
\begin{equation}
	\kappa_{\mathrm{des}}(s) = \begin{cases} 
		0.0817 s - 0.4087 		& \text{for}\: \ 5     \leq s \leq 7.6702\,,   \\
		0.8453 - 0.0817 s    	& \text{for}\: \ 7.6702 \leq s \leq 13.011\,,   \\
		0.0817 s - 1.2819 		& \text{for}\: \ 13.011 \leq s \leq 15.681\,,   \\
	\end{cases}
\end{equation}
according to \eqref{eq:kappaDesTurn}, \eqref{eq:kappaDesTurn2}.
Note that these numerical values are shown up to 4 digits only.

The final part is a ${\Delta s=5}$\,m long straight line parallel to the $x$~axis without changing the velocity.

	\subsection{Lateral and longitudinal control design}\label{sec:control}
	
	To simplify the control design, we assume that the desired maneuver can be carried out as a perturbation of the steady state of straight rolling with constant speed. 
	
	Linearizing the equations of motion \eqref{eq:1stOrderEQS} around the straight rolling steady state (i.e., ${\kappa(s)\equiv 0}$) with constant pitch rate $\dot\varphi_{*}$ enables us to decompose the lateral and longitudinal dynamics:
	\begin{equation}\label{eq:linearsys_all}\arraycolsep=2pt
		\begin{split}
			\begin{bmatrix}
				\dot{\mathbf x}_{\mathrm{lat}}
				\\
				\dot{\mathbf x}_{\mathrm{lon}}
			\end{bmatrix}
			&=\begin{bmatrix}
				\mathbf A_{\mathrm{lat}} & \mathbf 0\\
				\mathbf 0 & \mathbf A_{\mathrm{lon}}
			\end{bmatrix}	\begin{bmatrix}
				{\mathbf x}_{\mathrm{lat}}
				\\
				{\mathbf x}_{\mathrm{lon}}
			\end{bmatrix} + \begin{bmatrix}
				\mathbf B_{\mathrm{lat}} & \mathbf 0\\
				\mathbf 0 & \mathbf B_{\mathrm{lon}}
			\end{bmatrix}	\begin{bmatrix}
				{u}_{\mathrm{lat}}
				\\
				{u}_{\mathrm{lon}}
			\end{bmatrix}\!,
		\end{split}
	\end{equation}
	where the states and control inputs related to the lateral and longitudinal subsystems are
	\begin{equation}\arraycolsep=1pt\label{eq:latlon}
		\begin{array}{rlrl}
			{\mathbf x}_{\mathrm{lat}} &= \big[
			\omega_{1}
			\ \ \ \sigma_r
			\ \ \ r
			\ \ \ \vartheta
			\ \ \ \omega_{3}
			\ \ \ \chi
			\ \ \ \varepsilon
			\big]^{\mathsf T}, 
			&\qquad 
			{u}_{\mathrm{lat}} &= F,
			\\	
			{\mathbf x}_{\mathrm{lon}} &= \big[
			\tilde\omega_{2}
			\ \ \ \tilde\sigma_{\gamma}
			\ \ \ \gamma
			\ \ \ \tilde\varphi
			\ \ \ \tilde s
			\big]^{\mathsf T},
			&\qquad
			{u}_{\mathrm{lon}} &= T,
		\end{array}
	\end{equation}
	and
	\begin{equation}
		\begin{split}
			\tilde\omega_{2}      &= \omega_{2}       - \dot\varphi_{*},
			\\ 
            \tilde\sigma_{\gamma} &= \sigma_{\gamma}  - \dot\varphi_{*}R,
			\\
			\tilde\varphi         &= \varphi          - (\dot\varphi_{*}t + \varphi_{0}),
			\\ 
            \tilde s              &= s                - (\dot\varphi_{*}Rt + s_{0}).
		\end{split}
	\end{equation}
	
	The state and input matrices of the lateral subsystem are
	\begin{equation}\label{eq:A_LAT}\arraycolsep=2pt
		\mathbf A_{\mathrm{lat}} = \left[\begin{matrix}
			0      & 0 & a_{13} & a_{14} & a_{15} & 0      & 0 \\
			0      & 0 & 0      & a_{24} & a_{25} & 0      & 0 \\
			a_{31} & 1 & 0      & 0      & 0      & 0      & 0 \\
			1      & 0 & 0      & 0      & 0      & 0      & 0 \\
			a_{51} & 0 & 0      & 0      & 0      & 0      & 0 \\
			0      & 0 & 0      & 0      & 1      & 0      & 0 \\
			0      & 0 & 0      & 0      & 0      & a_{76} & 0 
		\end{matrix}\right]\!,
		\quad
		\mathbf B_{\mathrm{lat}} = \left[\begin{matrix}b_{11}\\b_{21}\\0\\0\\0\\0\\0\end{matrix}\right],%
	\end{equation}
	where
	\begin{equation}\arraycolsep=1em
		\begin{array}{lll}
			a_{13} = - {4 m_{1} g}/{c_1}, & \multicolumn{2}{l}{a_{14} = {4 g \big(m R + m_{2} (R + h)\big)}/{c_1},}
			\\
			\multicolumn{2}{l}{a_{15} = {2 R \dot\varphi_{*} \big(3 m R + 2 m_{2} (R + h)\big)}/{c_1},}
			&a_{24} = - g ,\\
			a_{25} = - R \dot\varphi_{*}  ,\qquad
			&a_{31} = R ,
			&a_{51} = -2 \dot\varphi_{*} , \\
			a_{76} = R\dot\varphi_{*} ,
			&b_{11} = -{4 R}/{c_1} ,
			&b_{21} = -1/m_1 ,
			\\
			\multicolumn{3}{c}{
				c_1 = 5 m R^{2} + 4 m_{2} (R + h)^{2}  .
			}
		\end{array}%
	\end{equation}
	Observe that apart from the system parameters, the lateral subsystem depends on the steady state pitch rate $\dot\varphi_{*}$.

	Note that at zero speed ${\dot\varphi^* = 0}$, 
	several terms in $\mathbf A_{\mathrm{lat}}$ disappear.
	As $a_{76}$ becomes zero, the lateral error state $\varepsilon$ cannot be influenced in the linearized model. 
	In this paper, we avoid this singularity by simply starting the simulations from a small nonzero initial longitudinal velocity.
	
	The characteristic polynomial of the lateral dynamics is
	\begin{equation}
		\left(\lambda^{4} + a_2\lambda^{2} +a_0\right) \lambda^{3} = 0,
	\end{equation}
	where the coefficients are 
	\begin{equation}
		\begin{split}
			a_0 &=  - a_{13} (a_{24}  + a_{25} a_{51}) ,
			\\
			a_2 &= - a_{13} a_{31} - a_{14} - a_{15} a_{51} ,
		\end{split}
	\end{equation}
	yielding the nonzero characteristic roots 
	\begin{equation}\label{eq:nonzeroroots}
		\lambda_{1,2,3,4} = \pm\sqrt{\dfrac{-a_2\pm\sqrt{a_2^2 - 4a_0}}{2}}.
	\end{equation}
	
\begin{figure}[!t]
		\includegraphics[width=\columnwidth]{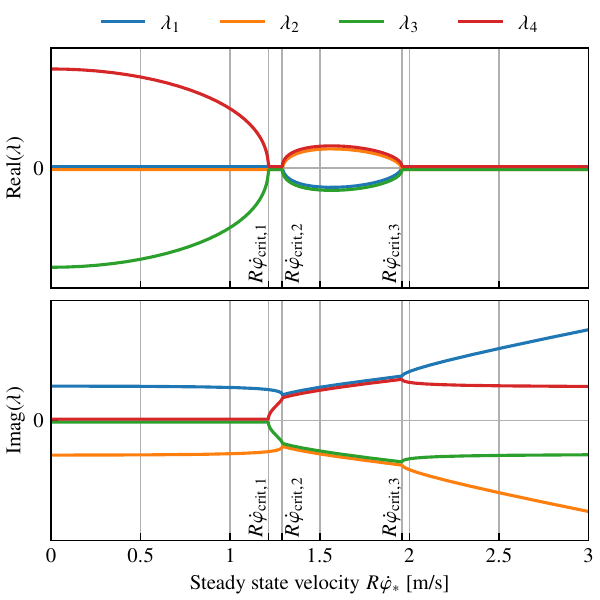}%
		\caption{Characteristic roots and critical velocities for the lateral dynamics, the critical velocities are approximately 1.21, 1.29, and 1.95 m/s.
		}
		\label{fig:crispeed}
	\end{figure}

	When ${a_0>0}$, ${a_2>0}$ and ${a_2^2 - 4a_0>0}$ these roots are along the imaginary axis and steady state motion is neutrally stable. 
	While ${a_2>0}$ always holds, the condition ${a_0>0}$ is equivalent to 
	\begin{equation}\label{eq:crit1}
		\dot\varphi_{*} > \dot{\varphi}_{*\mathrm{crit},1} = \sqrt{\frac{g}{2R}} .
	\end{equation}
	Note that the critical angular velocity $\dot\varphi_{\mathrm{crit},1}$ only depends on $R$ and $g$, it does not depend on the masses $m$, $m_1$, $m_2$ and the length $h$.
	One may show that 
	\begin{equation}
		a_2^2 - 4a_0 = b_4 \dot\varphi_{*}^4 +b_2 \dot\varphi_{*}^2+b_0 
	\end{equation}
	with ${b_4>0}$, ${b_2<0}$ and ${b_0>0}$.
	If ${b_2^2-4b_0b_4>0}$ holds then the condition ${a_2^2 - 4a_0>0}$ yields 
	\begin{equation}\label{eq:crit23}
		\begin{split}
			\dot\varphi_{*} < \dot\varphi_{\mathrm{crit},2}\,, 
			\quad {\rm or} \quad
			\dot\varphi_{*} > \dot\varphi_{\mathrm{crit},3}\,,
			\\
			\dot\varphi_{\mathrm{crit},2,3} = \sqrt{\dfrac{-b_2\mp\sqrt{b_2^2-4b_0b_4}}{2b_4}}\,.
		\end{split}
	\end{equation}
	In other words, the lateral dynamics is unstable for slow velocities ${\dot\varphi_{*}<\dot\varphi_{\mathrm{crit},1}}$ (in which case the unicycle topples), and for  ${\dot\varphi_{*} \in (\dot\varphi_{\mathrm{crit},2},\dot\varphi_{\mathrm{crit},3})}$ (in which case growing oscillations appear).
	This is illustrated in Figure~\ref{fig:crispeed}, where real and imaginary values of the roots \eqref{eq:nonzeroroots} are plotted as function of the speed $R\dot\varphi_{*}$ of the wheel center for the parameters in Table~\ref{tab:mechParams}.
	Indeed there exist roots with positive real parts when $R\dot\varphi_{*}$ is below $R\dot\varphi_{\mathrm{crit},1}$ and when it is between $R\dot\varphi_{\mathrm{crit},2}$ and $R\dot\varphi_{\mathrm{crit},3}$.

	\begin{table}[!t]
		\centering%
		\setlength{\tabcolsep}{1pt}%
		\caption{Parameter values used in the paper}\label{tab:mechParams}%
		\begin{tabular}{lcrl}
			\multicolumn{1}{c}{Parameter name} & \quad~ &   \multicolumn{2}{c}{Value~~~~~~}     \\ \hline
			Wheel mass                         &         &   $m$ & ${ = 4\,\mathrm{kg}      }$  \\
			Lateral actuator mass              &         & $m_1$ & ${ = 10\,\mathrm{kg}      }$  \\
			Longitudinal actuator mass         &         & $m_2$ & ${ = 10\,\mathrm{kg}      }$  \\
			Longitudinal actuator length       &         &   $h$ & ${ = 0.3\,\mathrm{m}      }$  \\
			Wheel radius                       &         &   $R$ & ${ = 0.3\,\mathrm{m}      }$  \\
			\hline
		\end{tabular}
	\end{table}
		
	The longitudinal subsystem has the state and input matrices
	\begin{equation}\arraycolsep=2pt
		\begin{split}
			\mathbf A_{\mathrm{lon}} &= \left[\begin{matrix}0 & 0 & a_{8\mathrm A} & 0 & 0\\0 & 0 & a_{9\mathrm A} & 0 & 0\\a_{\mathrm A8} & a_{\mathrm A9} & 0 & 0 & 0\\1 & 0 & 0 & 0 & 0\\a_{\mathrm C8} & 0 & 0 & 0 & 0\end{matrix}\right]\!, 
			\qquad
			\mathbf B_{\mathrm{lon}}  = \left[\begin{matrix}b_{82}\\b_{92}\\0\\0\\0\end{matrix}\right]%
			\!,
		\end{split}
	\end{equation}
	where the indices of the matrix elements refer to the indices in the full system \eqref{eq:linearsys_all} with $\mathrm A$ and $\mathrm C$ referring to the 10\textsuperscript{th} and 12\textsuperscript{th} indices, respectively. 
	The matrix elements are
	\begin{equation}\arraycolsep=6pt
		\begin{array}{lll}
			\multicolumn{2}{l}{a_{8\mathrm A} = - {2 m_{2} g}/{\big(3 m R + 2 m_{1} R \big)},}
			&a_{9\mathrm A} = g,
			\\
			a_{\mathrm A8} = - {R}/{h},\quad~~~
			&a_{\mathrm A9} = {1}/{h},
			&a_{\mathrm C8} = {R},
			\\
			\multicolumn{2}{l}{\,b_{82} = {2 \big(R + h\big)}/{\big((3 m  + 2 m_{1})R^{2} h \big)},}
			&b_{92}\, = -{1}/\big({m_{2} h}\big) ,
		\end{array}%
	\end{equation}
	which do not depend on the steady state pitch rate $\dot\varphi_{*}$.
	
	The corresponding characteristic polynomial
	\begin{equation}
		\left(\lambda^{2} - a_{\mathrm A8} a_{8\mathrm A} - a_{\mathrm A9} a_{9\mathrm A}\right) \lambda^{3} = 0\,,
	\end{equation}
	yields the nonzero characteristic roots
	\begin{equation}
		\lambda_{8,9} = \pm \sqrt{a_{\mathrm A8} a_{8\mathrm A} + a_{\mathrm A9} a_{9\mathrm A} 
		}= \pm\sqrt{\frac{({3 m + 2 m_{1} + 2 m_{2}}){g}}{ ({3 m + 2 m_{1}}){h}}} . 
	\end{equation}
	Since $\lambda_{8,9}\in\mathbb{R}$ and $\lambda_{8} > 0$ the longitudinal subsystem is always unstable without applying control.

	One can conclude that neither the lateral nor the longitudinal subsystems are state controllable as the controllability matrices
	\begin{equation}\setlength{\arraycolsep}{1ex}
		\begin{split}
			\bm{\mathcal{M}}_{l} &= \begin{bmatrix}
				\mathbf B_l 
				& \mathbf A_l \mathbf B_l
				& \mathbf A_l^2 \mathbf B_l
				& \dots
				&
			\end{bmatrix}\,
		\end{split}
	\end{equation} 
	do not have full rank; 
	here the subscript $l$ stands for either `$\mathrm{lat}$' or `$\mathrm{lon}$' referring to the corresponding subsystem.
	However, choosing the control outputs to be
	\begin{equation}\label{eq:importantStates}\setlength{\arraycolsep}{1ex}
		\begin{split}
			\mathbf y_{\mathrm{lat}} &= \mathbf{C}_{\mathrm{lat}}\mathbf x_{\mathrm{lat}} := \begin{bmatrix}
				\omega_{1} 
				&  \sigma_r
				&  r 
				&  \vartheta 
                &  \chi 
				&  \varepsilon
			\end{bmatrix}^{\mathsf T}\!, \\
			\mathbf y_{\mathrm{lon}} &= \mathbf{C}_{\mathrm{lon}}\mathbf x_{\mathrm{lon}} := \begin{bmatrix}
				\tilde{\omega}_{2} 
				& \tilde{\sigma}_{\gamma}
				& \gamma 
				& \tilde{s}
			\end{bmatrix}^{\mathsf T}\!,
		\end{split}
	\end{equation} 
	cf.~\eqref{eq:latlon}, both subsystems are output controllable since the output controllability matrices 
	\begin{equation}\label{eq:controllabilityMTRXS}\setlength{\arraycolsep}{1ex}
		\begin{split}
			\bm{\mathcal{M}}_{\mathrm{oc},l} &= \begin{bmatrix}
				\mathbf C_l \mathbf B_l 
				& \mathbf C_l \mathbf A_l \mathbf B_l
				& \mathbf C_l \mathbf A_l^2 \mathbf B_l
				& \dots
				&
			\end{bmatrix}\,
		\end{split}
	\end{equation} 
	have full rank.

	We consider the linear output feedback with control laws%
    \begin{equation}\label{eq:control_input}
		\begin{split}
			F &:= D_\vartheta \omega_{1} 
			+  D_r \sigma_r
			+  P_{r}r 
			+  P_\vartheta \vartheta 
			+  P_{\chi}\chi
			+  P_{\varepsilon}\varepsilon \,,
			\\
			T &:= D_\varphi \hat{\omega}_{2} 
			+ D_\gamma \hat{\sigma}_{\gamma} 
			+ P_\gamma \gamma 
			+ P_s \hat{ s}\,,
		\end{split}
	\end{equation}
	where the error signals in case of nonzero desired values are denoted by hats:
	\begin{equation}
    \begin{split}
		\hat\omega_{2} &= \omega_{2} -\omega_{2,\mathrm{des}} ,
        \\
		\hat\sigma_{\gamma} &= \sigma_{\gamma} -\sigma_{\gamma ,\mathrm{des}} ,
        \\
		\hat s &= s - s_{\mathrm{des}}.
    \end{split}
	\end{equation}
	The path planning process provides the desired arc length $s_{\mathrm{des}}$ and velocity profile ${v_{\mathrm{des}} = \dot s_{\mathrm{des}}}$ while the desired pseudovelocities can be obtained as
	\begin{equation}
		\begin{split}
			\omega_{2,\mathrm{des}}  &=\dot s_{\mathrm{des}}/R,
			\\
			\sigma_{\gamma ,\mathrm{des}} &= \omega_{2,\mathrm{des}}  R \cos{\gamma} + \omega_{3} h  \tan{\vartheta}.
		\end{split}
	\end{equation}  
	Here the desired velocity $\sigma_{\gamma ,\mathrm{des}}$ of $m_2$ considers zero pitch rate ${\dot\gamma =0}$, while compensating for nonzero tilt angle $\vartheta$, non-zero pitch angle ${\gamma}$, and nonzero yaw rate ${\dot\psi}$ during the turning sections.

	\section{Simulations}\label{sec:simulation}

	In order to test the performance of the integrated path planning and control design, the closed-loop nonlinear behavior of the unicycle is simulated for a set of family of lane change maneuvers.
	Three different desired speed values are considered: 
	1\,m/s which is below the first critical speed $R\dot\varphi_{\mathrm{crit},1}$ given in \eqref{eq:crit1}, 
	1.5\,m/s which is between the second and third critical speeds $R\dot\varphi_{\mathrm{crit},2}$ and $R\dot\varphi_{\mathrm{crit},3}$ given in \eqref{eq:crit23}, 
	and 3\,m/s which is above the third critical speed $R\dot\varphi_{\mathrm{crit},3}$; see Figure~\ref{fig:crispeed}.
	The parameters listed in Table~\ref{tab:mechParams} are used for the simulations.
	
	The linear output feedback \eqref{eq:control_input} is applied to the fully nonlinear equations of motion \eqref{eq:dsigmaDynEQS}, \eqref{eq:kinematicEQS_withPATH}.
	The control gains are tuned such that the assignable ten characteristic roots are set to ${\lambda_i = -12\,\mathrm{s}^{-1}}$, and there are two additional  roots at zero corresponding to the two uncontrollable modes.

    The initial conditions for all simulations are as follows: ${\sigma_\gamma(0) = 0.001\, \mathrm{m/s}}$,
    ${\sigma_2(0) = \sigma_\gamma(0)/R \,\mathrm{rad/s}}$, the rest of the states initialized to be zeros.
	Following the path plan in Section~\ref{sec:Maneuver}, the lane change maneuver consists of three parts: 
	(I) the initial accelerating straight part, 
	(II) the lane change with constant velocity, 
	and (III) the final straight part with constant velocity.
	For part (I), length ${\Delta s_{\mathrm I} = 5}$\,m is considered for accelerating while ${\kappa_{\mathrm{I,des}}(s)\equiv0\,\mathrm{m}^{-1}}$.
	The arc length and speed profiles ${s_{\mathrm{I},\mathrm{des}}(t)}$ and ${v_{\mathrm{I},\mathrm{des}}(t)}$ are obtained from \eqref{eq:sdesdesign} and \eqref{eq:vdesdesign}, respectively,
	while using ${\Delta v_{\mathrm{I}}=1,1.5}$ and $3$\,m/s for the three different speeds considered.
	For the turning part, a family of three-clothoid trajectories is designed with clothoid section ratios~${\mathcal S \in\big[0.05,0.8\big]}$ while using ${\Delta x=10}$\,m, ${\Delta y=3}$\,m, ${\Delta \psi=0}$.
	The curvature profile $\kappa_{\mathrm{II,des}}(s)$ is obtained according to \eqref{eq:kappaDesTurn}, \eqref{eq:kappaDesTurn2}.
	The arc length $\Delta s_{\mathrm{II}}$ of the lane change part depends on the clothoid segment ratio~$\mathcal S$.
	The desired velocity during the lane change is set to be the instantaneous velocity.
	The final part is a ${\Delta s_{\mathrm{III}}=5}$\,m long straight segment (${\kappa_{\mathrm{III,des}}(s)\equiv0\,\mathrm{m}^{-1}}$) where the velocity is kept constant at the final value of the previous part.
	
\begin{figure*}[!p]
		\centering
		\includegraphics[width=0.99\textwidth]{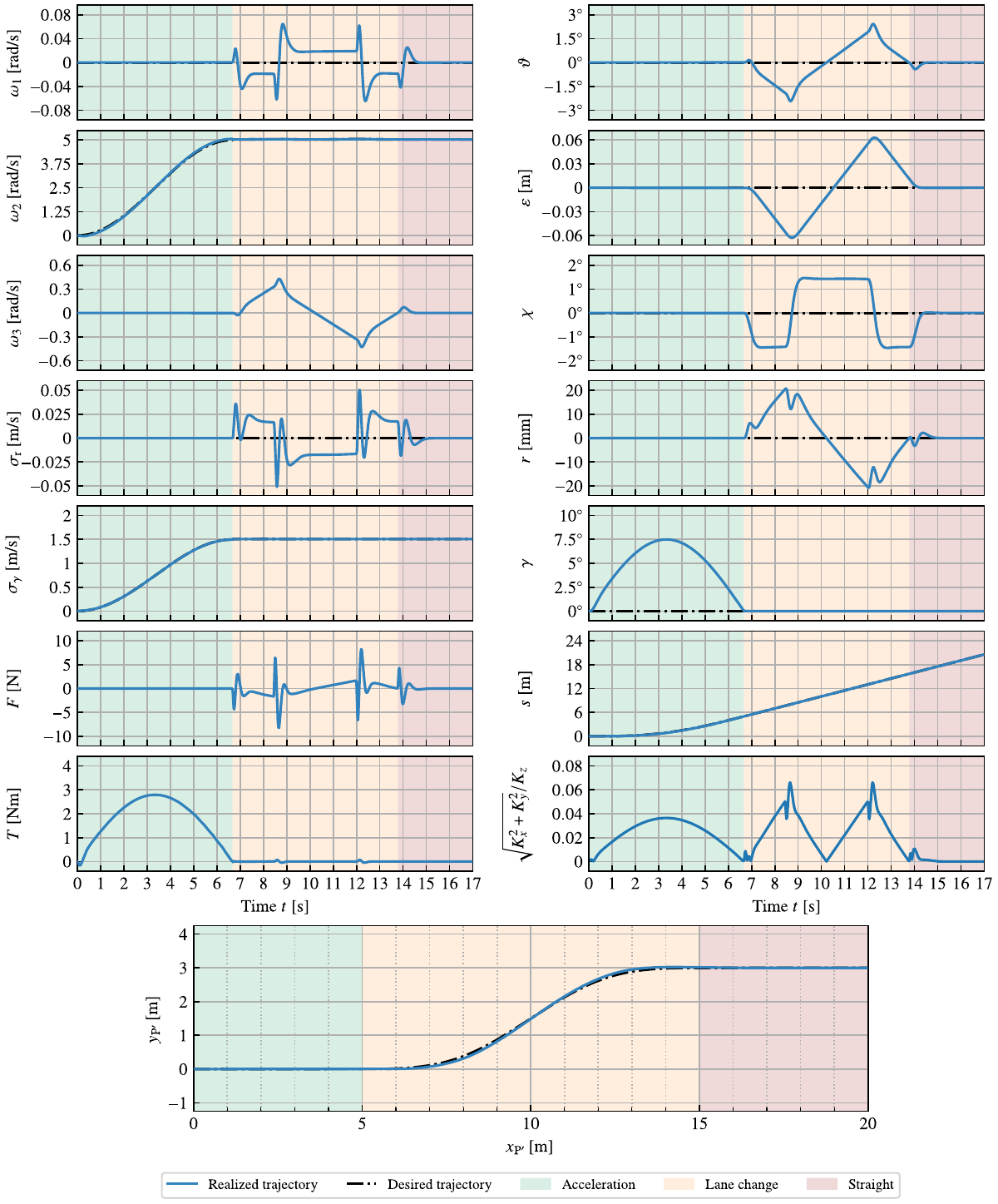}
		\caption{Simulation of the lane change maneuver with the robotic unicycle, for ${R\dot\varphi_{\mathrm{des}}=1.5}$\,m/s, ${\Delta x = 10}$\,m, ${\Delta y=3}$\,m and ${\mathcal{S}=0.5}$}
		\label{fig:simulation_S05}
	\end{figure*}

	The numerical simulations show that the unicycle is able to carry out various lane change maneuvers with the unicycle model \eqref{eq:dsigmaDynEQS}, \eqref{eq:kinematicEQS_withPATH} and control law \eqref{eq:control_input}.
	The time evolution of the states, the control inputs, and contact force components are shown in Figure~\ref{fig:simulation_S05} 
	for the longitudinal speed 1.5\,m/s.
	The maneuver parts are distinguished with different background colors: green is used for the initial accelerating part (I), yellow for the lane change part (II), and red for the final straight part (III).
	The simulated trajectories are shown by the blue curves, while the dashed black curves indicate the desired states.
	The longitudinal actuator with torque $T$ is utilized to increase the speed to the desired value during the acceleration phase, however, it is also applied during the turning phase for balancing the pendulum.
	The lateral actuator with force $F$ is used in the turning phase to steer the unicycle when following the curved reference path while also keeping the unicycle near the vertical position.
	After the lane change the unicycle follows the reference straight path in a stable manner.
	
	The powers $P_F$ and $P_T$ of the control inputs $F$ and $T$ can be calculated as
	\begin{equation}
		\begin{split}
			P_F &= \mathbf F \cdot \mathbf v_{\mathrm{C}} + (-\mathbf F)\cdot \mathbf v_{\!\mathrm{A}} 
			= -\big(\omega_{1}  R + \sigma_r\big) F \,, 
			\\
			P_T &= \mathbf{T} \cdot \bm \omega + (-\mathbf T)\cdot \bm \Omega 
			= \frac{1}{h}\big(\omega_{2}({R \cos{\gamma}}+ h)  - {\sigma_{\gamma}}\big)T \,,
		\end{split}
	\end{equation}
	respectively, where the velocities $\mathbf v_{\mathrm{C}}$ and $\mathbf v_{\!\mathrm{A}}$ are given by \eqref{eq:angularaccel_disc} and \eqref{eq:vA_aA}, respectively, while the angular velocity $\bm \Omega$ of the pendulum is given by \eqref{eq:pendangvel}.
	The required maximal powers ${|P_F|_\mathrm{max}\approx 1}$\,W and ${|P_T|_\mathrm{max} \approx 10}$\,W are also well within the realistic range for real world actuators.
	
\begin{figure*}
	\centering
	\includegraphics[width=0.999\textwidth]{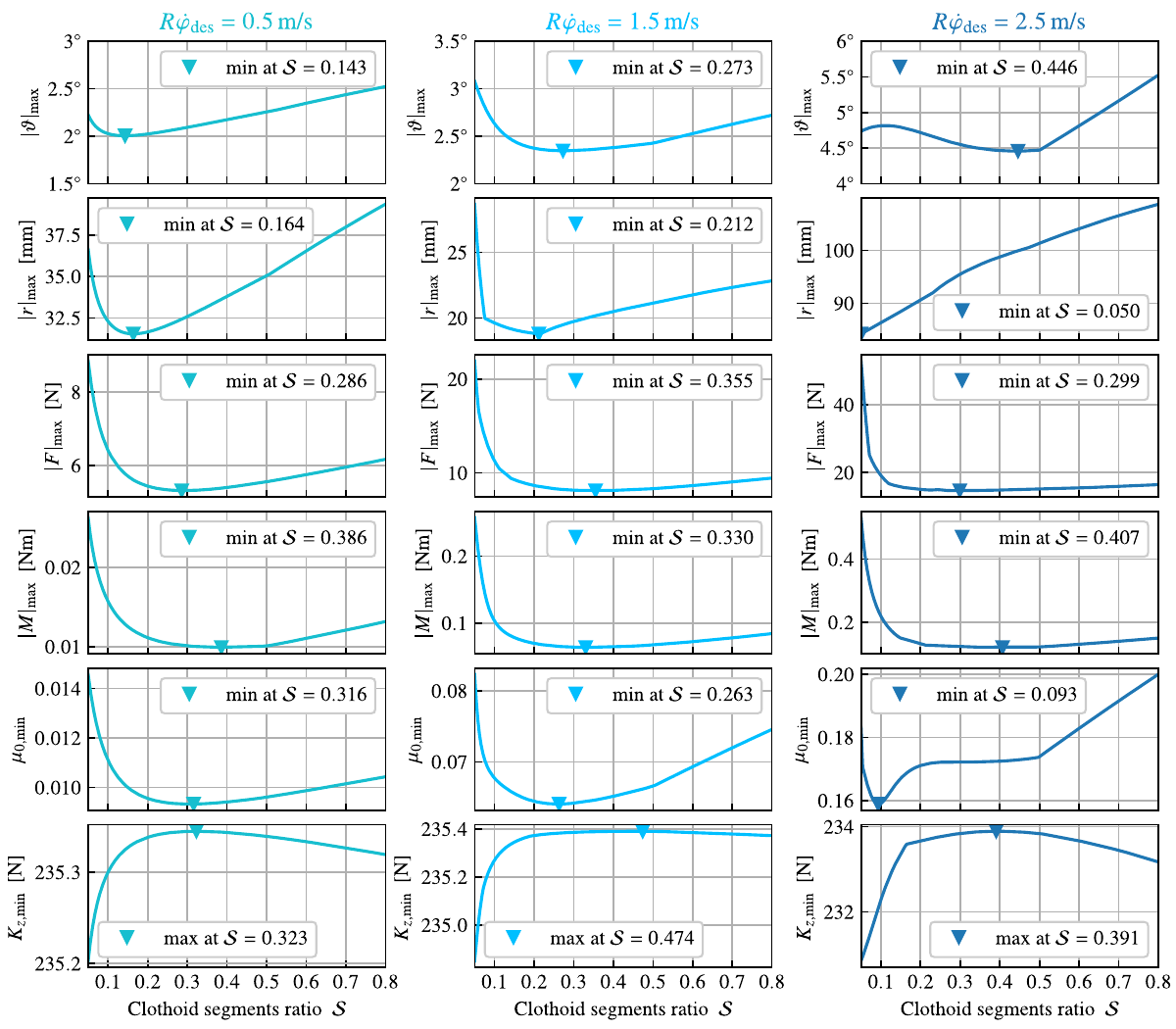}   
	\caption{Maximal tilt angles, maximal lateral actuator position, maximal control inputs, the necessary static friction coefficient needed to avoid slipping of the wheel, and the minimal vertical contact force as a function of the clothoid segment ratio ${\mathcal S\in\big[0.05,0.8\big]}$ for longitudinal speeds $R\dot\varphi_{\mathrm{des}}=1, 1.5$ and $3$\,m/s.
	}
	\label{fig:Maximal_wrt_Ratios}
\end{figure*}

    The effects of the desired path (i.e.~the choice of the clothoid segment ratio $\mathcal S$) on the maximal tilt angle, on the maximal point mass position, on the maximal actuator force and torque, and on the required static friction coefficient along the maneuvers are shown in Figure~\ref{fig:Maximal_wrt_Ratios}.
	Observe that the minima of these quantities vary significantly with the longitudinal speed that is different in each column.
	Designing the lane change maneuver, by choosing an appropriate clothoid segment ratio, is therefore crucial for achieving good control performance.
	Poorly choosing the clothoid segment ratio can lead to large tilt angles (increasing the chance of tipping over) and large lateral mass position (requiring long axle).
	This can significantly increase (even double or triple) the necessary control input values and may result in slipping of the wheel on surfaces with lower friction coefficients.

\newpage
~
\newpage
~
\newpage

\section{Conclusion}\label{sec:conclusion}

In this paper, we presented an integrated approach to modeling, path planning, and path following for a robotic unicycle capable of acceleration and turning maneuvers using two actuators. 
The full nonlinear equations of motion and the wheel-ground contact force components were derived in closed form. 
A path-following transformation was introduced so that the longitudinal, lateral, and angular errors relative to the reference path become system states, making them well suited for feedback control. 
Motion planning was performed using segmented paths composed of straight sections and clothoid-based curves. 
By linearizing the dynamics along a straight path, the lateral and longitudinal dynamics were decoupled, and linear state-feedback controllers were designed for the resulting subsystems. 
The closed-loop performance was evaluated through numerical simulations by applying the controllers to the full nonlinear governing equations. 
The results showed that path planning significantly affects the resulting motion and that paths can be optimized based on closed-loop performance.

Future work will improve the path design by incorporating physically consistent desired profiles for the tilt angle and point-mass position along curved sections, as well as desired pendulum-angle profiles along straight sections. 
Future efforts will also incorporate turning-rolling steady states into the control design. 
In addition, we will extend the model to include more realistic tire-ground contact effects, such as rolling resistance, lateral deformation, partial slip, and transitions between sticking and sliding, which are expected to play an important role during aggressive maneuvers. 
These extensions will move the proposed framework toward experimentally validated, agile robotic unicycle locomotion.

	\appendix

	\section{Derivation of governing equations}
	\label{app:govEqn}
	A vector quantity $\VECTOR_{\mathcal{F}_j}$ or a tensor quantity $\TENSOR_{\mathcal{F}_j}$ given in frame $\mathcal{F}_j$ can be transformed in an other frame $\mathcal{F}_i$ as 
	\begin{equation}
		\VECTOR_{\mathcal{F}_i} = \mathbf T_{ij}\,\VECTOR_{\mathcal{F}_j},\qquad
		\TENSOR_{\mathcal{F}_i} = \mathbf T_{ij}\,\TENSOR_{\mathcal{F}_j}\mathbf T_{ji},
	\end{equation}
	where the transformation matrices corresponding to the coordinate frames introduced in Section~\ref{sec:mechmodel} are
	\begin{align}
		\mathbf T_{01} &= \left[\arraycolsep=4pt
		\begin{matrix}
			\cos{\psi_{}\,\!} & - \sin{\psi_{}\,\!} & 0\\
			\sin{\psi_{}\,\!} & \cos{\psi_{}\,\!} & 0\\
			0 & 0 & 1
		\end{matrix}\right]\!,
		\nonumber\\
			\mathbf T_{12} &= \left[\arraycolsep=4pt
			\begin{matrix}
				1 & 0 & 0\\
				0 & \cos{\vartheta_{}\,\!} & - \sin{\vartheta_{}\,\!}\\
				0 & \sin{\vartheta_{}\,\!}    & \cos{\vartheta_{}\,\!}
			\end{matrix}\right]\!,
		\\
		\mathbf T_{23} &= \left[\arraycolsep=4.5pt
		\begin{matrix}
			\cos{\gamma_{}\,\!} & 0 & \sin{\gamma_{}\,\!}\\
			0 & 1 & 0\\
			- \sin{\gamma_{}\,\!} & 0 & \cos{\gamma_{}\,\!}
		\end{matrix}\right]\!,
		\nonumber\\
			\mathbf T_{02} &= \mathbf T_{01} \mathbf T_{12},
            \nonumber
            \\
			\mathbf T_{03} &= \mathbf T_{01} \mathbf T_{12}\mathbf T_{23}, 
			\nonumber
	\end{align}
where ${\mathbf T_{ji} = \mathbf T_{ij}^{-1} = \mathbf T_{ij}^{\top}}$.
	
	The Appellian approach~\citep{Appell_1900, Gibbs_1879, qin2022nonholonomic} assumes the equation of motion in the form 
	\begin{equation}\label{eq:appell}\arraycolsep=1pt
		\begin{array}{rll}
			\dfrac{\partial S}{\partial \dot\sigma_{\!j}} &= \Pi_j, & j=1,\dots,n_\sigma\,,  
			\\[1em]
			\dot q_k &= \displaystyle\sum_{j=1}^{n_\sigma} f_{kj} \sigma_j,\quad~ & k=1,\dots,n_q\,,
		\end{array}
	\end{equation}
	where $S$ is the the acceleration energy, $\Pi_j$ is a pseudoforce.
	The acceleration energy 
	\begin{equation}
		\label{eq:acc_energy_disc}
		\begin{split}
			S = S_\mathrm{w} + S_{m_1} + S_{m_2},
		\end{split}
	\end{equation}
	of the unicycle is composed of three parts related to wheel and the point masses:
	\begin{equation}\label{eq:accel_energy_wheel}
		\begin{split}
			S_\mathrm{\!w} &= \frac{1}{2} m \mathbf a_{\rm C}^2 + \frac{1}{2}\bm\alpha \cdot \mathbf J_{\rm C} \bm\alpha + \bm\alpha \cdot ( \bm\omega \times \mathbf J_{\rm C} \bm\omega )\,,
            \\
			S_{\!m_1} &= \frac{1}{2} m_1 \mathbf a_{\rm A}^2,
			\\
			S_{\!m_2} &= \frac{1}{2} m_2 \mathbf a_{\rm B}^2.
		\end{split}
	\end{equation}
	The velocity and acceleration of the wheel center, the angular velocity and angular acceleration of the wheel, and the mass moment of inertia of the wheel are
	\begin{align}\label{eq:angularaccel_disc}
		\begin{split}
        \mathbf v_{\mathrm C} &=		
        \begin{bmatrix}
		R \omega_{2} \\
		- R \omega_{1} \\
		0
		\end{bmatrix}_{{\mathcal F}_{2}}\!\!, 
        \quad
			\mathbf a_{\rm C} = 
			\begin{bmatrix}
				R (\dot \omega_{2} + \omega_{1} \omega_{3})   \\
				- R (\dot \omega_{1} - \omega_{2} \omega_{3}) \\
				- R (\omega_{1}^{2} + \omega_{2} \omega_{3} \tan{\vartheta})
			\end{bmatrix}_{{\mathcal F}_{2}}\!\!,
            \\
			\bm\omega &= 
			\begin{bmatrix}
				\omega_{1}\\
				\omega_{2}\\
				\omega_{3}
			\end{bmatrix}_{{\mathcal F}_{2}}\!\!,
            \quad\,\,\,
            \bm\alpha = 
			\begin{bmatrix}
				\dot \omega_{1} - \omega_{2} \omega_{3} + \omega_{3}^{2} \tan{\vartheta}\\
				\dot \omega_{2}\\
				\dot \omega_{3} + \omega_{1} \omega_{2} - \omega_{1} \omega_{3} \tan{\vartheta}
			\end{bmatrix}_{{\mathcal F}_{2}}\!\!,
            \\
			\mathbf J_{\rm C} &= \frac{mR^2}{4}
			\begin{bmatrix}
				1 & 0 & 0\\
				0 & 2 & 0\\
				0 & 0 & 1
			\end{bmatrix}_{{\mathcal F}_{2}}\!\!.
			\end{split}
	\end{align}
	The velocity and the acceleration of $m_1$ are
	\begin{equation}\label{eq:vA_aA}
		\begin{split}
			\mathbf v_{\!\rm A} &= 
			\begin{bmatrix}
				\omega_{2} R  - \omega_{3} r
				\\ 
				\sigma_r
				\\ 
				\omega_{1} r
			\end{bmatrix}_{{\mathcal F}_{2}},
			\\
			\mathbf a_{\rm A} &= \begin{bmatrix}
				\dot \omega_{2}  R  - \dot \omega_{3} r  +  \omega_{1} \omega_{3}  \big(r \tan{\vartheta} - R\big)- 2 \omega_{3}\sigma_r
				\\
				\dot \sigma_r- \omega_{1}^{2} r - \omega_{3}^{2} r  + \omega_{2} \omega_{3} R \ \ \ 
				\\
				\dot \omega_{1} r + \omega_{1}^{2} R  + 2 \omega_{1}  \sigma_r + (\omega_{3}^{2} r - \omega_{2} \omega_{3} R  )\tan{\vartheta} 
			\end{bmatrix}_{{\mathcal F}_{2}} \!\!,
		\end{split}
	\end{equation} 
	while the velocity and the acceleration of $m_2$ are
	\begin{equation}\label{eq:vB_aB}
		\begin{split}
			\mathbf v_{\!\rm B} &= 
			\begin{bmatrix}
				\sigma_{\gamma}
				\\
				\omega_{1} ( h  \cos{\gamma} - R)  + \omega_{3} h  \sin{\gamma}
				\\
				\omega_{2}R  \sin{\gamma}
			\end{bmatrix}%
			_{{\mathcal F}_{3}}\!\!,%
\\
			\mathbf a_{\rm B} &= \begin{bmatrix}
				a_{\mathrm{B}x} \\ a_{\mathrm{B}y} \\ a_{\mathrm{B}z}
			\end{bmatrix}_{{\mathcal F}_{3}}\!\!,
		\end{split}
	\end{equation}
	with
	\begin{equation} \label{eq:aB_componenets}
		\begin{split}
			a_{\mathrm{B}x} &=  \frac{1}{h}\big(
			\dot \sigma_{\gamma} h
			+ \omega_{1}^{2}(R  + h \cos{\gamma}) h \sin{\gamma}
			- \omega_{2}^{2}  R^{2} \sin{\gamma} \cos{\gamma} 
			\\&\quad  - \omega_{3}^{2} h^{2}  \sin{\gamma} \cos{\gamma} 
			+ \omega_{2} \sigma_{\gamma}  R \sin{\gamma}
			\\&\quad  + \omega_{1} \omega_{3} (R  \cos{\gamma} 
			+ 2 h \cos^{2}{\gamma} 
			- h) h 
			\big),\\
			a_{\mathrm{B}y} &=   
			- \dot \omega_{1} (R + h \cos{\gamma}) 
			+ \dot \omega_{3}  h \sin{\gamma} 
			- \omega_{3}^{2} h  \cos{\gamma} \tan{\vartheta} 
			\\&\quad  - 2 \omega_{1} \omega_{2}  R \sin{\gamma} \cos{\gamma} 
			-\omega_{1} \omega_{3} h  \sin{\gamma} \tan{\vartheta} 
			\\&\quad  + \omega_{2} \omega_{3} (1\!-\!2 \cos^{2}{\gamma}) R
			+ 2 \omega_{1} \sigma_{\gamma}  \sin{\gamma} 
			+ 2 \omega_{3} \sigma_{\gamma}  \cos{\gamma},
			\\
			a_{\mathrm{B}z} &=  \frac{1}{h}\big(
			\dot \omega_{2}  R h \sin{\gamma}
			-   \omega_{1}^{2} ( R +  h \cos{\gamma}) h \cos{\gamma}
			- \sigma_{\gamma}^{2}
			\\&\quad  
			- \omega_{2}^{2}  R^{2} \cos^{2}{\gamma} 
			-  \omega_{3}^{2}  \sin^{2}{\gamma} h^{2} - \omega_{2} \omega_{3} R h  \cos{\gamma} \tan{\vartheta} 
			\\&\quad 
			+ 2 R \sigma_{\gamma} \omega_{2} \cos{\gamma} 
			+  \omega_{1} \omega_{3}  (R \sin{\gamma} + 2  \sin{\gamma} \cos{\gamma}) h
			\big).
		\end{split}
	\end{equation}
	The acceleration energies of the corresponding parts are
	{\par\nobreak\vfil\penalty0\vfilneg\vtop\bgroup
	\begin{equation}
	\label{eq:acc_energy_wheel}
	\begin{split}
		S_\mathrm{\!w} &= \frac{m R^2}{8} \Big(
			5 \dot \omega_{1}^{2} + 6 \dot \omega_{2}^{2} + \dot \omega_{3}^{2} 
			+ (2 \omega_{3}^{2} \tan{\vartheta} - 12 \omega_{2} \omega_{3})\dot \omega_{1}
			\\&\quad
			+ 8 \omega_{1} \omega_{3} \dot \omega_{2}
			+ (4 \omega_{1} \omega_{2} - 2 \omega_{1} \omega_{3} \tan{\vartheta}) \dot \omega_{3} 
			\Big)
			+ \dots\,, 
    \end{split}
    \end{equation}%
    \par\xdef\tpd{\the\prevdepth}\egroup\prevdepth=\tpd}
    \newpage
    \begin{equation}
    \begin{split}
	    S_{\!m_1} &= \frac{m_{1} }{2}\Big(
			r^{2}  \dot \omega_{1}^{2} 
			+ R^{2}  \dot \omega_{2}^{2} 
			+ r^{2}  \dot \omega_{3}^{2} 
			+ \dot \sigma_{\mathrm{r}}^{2}
			- 2 R r \dot \omega_{2} \dot \omega_{3} 
			\\&\quad
			+ 2 \big(\omega_{1}^{2} R - \omega_{2} \omega_{3} R  \tan{\vartheta} + 2 \omega_{1} \sigma_{\mathrm{r}}  + \omega_{3}^{2} r \tan{\vartheta}\big) r \dot \omega_{1} 	
			\\&\quad
			- 2 R  \omega_{3} \big(\omega_{1} R + 2 \sigma_{\mathrm{r}} - \omega_{1} 
			r \tan{\vartheta}\big) \dot \omega_{2} 
			\\&\quad
			+ 2  \omega_{3} \big(\omega_{1} (R  - r \tan{\vartheta}) + 2 \sigma_{\mathrm{r}} \big)r \dot \omega_{3} 
			\\&\quad
			- 2 \big(\omega_{1}^{2} r + \omega_{3}^{2} r - R \omega_{2} \omega_{3}\big) \dot \sigma_{\mathrm{r}} 	
			\Big)
			+ \dots\,,
    \end{split}
    \end{equation}
    \begin{equation}
    \begin{split}
	    S_{\!m_2} &= \frac{m_{2}}{2 h}\Big(
			\big( R^{2} h + 2 R h^{2} \cos{\gamma} + h^{3} \cos^{2}{\gamma}\big)  \dot \omega_{1}^{2} 
			\\&\quad
			+ R^{2} h \sin^{2}{\gamma}  \dot \omega_{2}^{2} + h^{3} \sin^{2}{\gamma}  \dot \omega_{3}^{2} + h  \dot \sigma_{\gamma}^{2}
			\\&\quad
			- \big(	2 R h^{2} \sin{\gamma} + 2 h^{3} \sin{\gamma} \cos{\gamma}\big) \dot \omega_{1}  \dot \omega_{3} 
			\\&\quad
			+ \big(
			\omega_{3}^{2} ( 2 R h^{2} \cos{\gamma} \tan{\vartheta} + 2 h^{3} \cos^{2}{\gamma} \tan{\vartheta}) 
			\\&\qquad	
			+ 4 \omega_{1} \omega_{2} Rh( R  \cos{\gamma} + h \cos^{2}{\gamma} ) \sin{\gamma}
			\\&\qquad
			+ 2 \omega_{1} \omega_{3} (  R h^{2}  +  h^{3} \cos{\gamma} ) \sin{\gamma} \tan{\vartheta}
			\\&\qquad	
			- 4 \omega_{1}  \sigma_{\gamma} ( R h  
			+  h^{2}  \cos{\gamma}) \sin{\gamma}
			\\&\qquad	
			- 4 \omega_{3}  \sigma_{\gamma} ( R h \cos{\gamma} 
			+  h^{2} \cos^{2}{\gamma})
			\\&\qquad	
			+ 2Rh\omega_{2} \omega_{3} ( 2R \cos^{2}{\gamma} - R + 2h \cos^{3}{\gamma} 
			\\&\qquad\quad
			- h \cos{\gamma})
			\big) \dot \omega_{1} 
			\\&\quad
			- \big(
			\omega_{1}^{2} (2 R^{2} h \sin{\gamma} \cos{\gamma} - 2 R h^{2} \sin^{3}{\gamma} + 2 R h^{2} \sin{\gamma})
			\\&\qquad	
			+ 2 \omega_{2}^{2}  R^{3} \cos^{2}{\gamma} \sin{\gamma}
			+ 2 \omega_{3}^{2}R h^{2}  \sin^{3}{\gamma} 
			\\&\qquad	
			+ 2 \sigma_{\gamma}^{2} R  \sin{\gamma} 
			+2 \omega_{2} \omega_{3} R^{2} h \sin{\gamma} \cos{\gamma} \tan{\vartheta} 
			\\&\qquad	
			- 2 \omega_{1} \omega_{3} Rh( R  + 2h  \cos{\gamma}) \sin^{2}{\gamma}
			\\&\qquad	
			- 4 \omega_{2}  \sigma_{\gamma} R^{2} \sin{\gamma} \cos{\gamma}
			\big) \dot \omega_{2}
			\\
			&\quad
			- 2 \big(
			\omega_{3}^{2}  h^{3}\sin{\gamma} \cos{\gamma} \tan{\vartheta} 
			+ 2 \omega_{1} \omega_{2} R h^{2}  \sin^{2}{\gamma} \cos{\gamma} 
			\\&\qquad
			+  \omega_{1} \omega_{3} h^{3}  \sin^{2}{\gamma} \tan{\vartheta}  
			- 2  \omega_{1} \sigma_{\gamma} h^{2} \sin^{2}{\gamma}   
			\\&\qquad
			+  \omega_{2} \omega_{3}  R h^{2} (2  \cos^{2}{\gamma}  - 1 )\sin{\gamma}
			\\&\qquad
			- 2 \omega_{3} \sigma_{\gamma}  h^{2}  \sin{\gamma} \cos{\gamma}  
			\big) \dot \omega_{3} 
			\\
			&\quad
			+ \big(
			\omega_{1}^{2}  ( 2 R h \sin{\gamma} + 2 h^{2} \sin{\gamma} \cos{\gamma})
			\\&\qquad
			-2 \omega_{2}^{2}  R^{2} \sin{\gamma} \cos{\gamma} 
			- 2 \omega_{3}^{2} h^{2}  \sin{\gamma} \cos{\gamma} 
			\\&\qquad
			+ 2 \omega_{1} \omega_{3} (  R h \cos{\gamma} + 2 h^{2} \cos^{2}{\gamma} - h^{2}) 
			\\&\qquad
			+ 2 \omega_{2} \sigma_{\gamma} R  \sin{\gamma} 
			\big) \dot \sigma_{\gamma} 
			\Big)
			+ \dots\,,
		\end{split}
	\end{equation}
	where $\dots$ refer to further terms which are independent of the pseudo accelerations, so they do not appear in the equations of motion.

	The pseudoforces can be expressed using the virtual power 
	\begin{equation}
		\begin{split}
			\delta P &= \mathbf G \cdot \delta \mathbf v_{\rm C} + \mathbf G_{\!\rm A} \cdot \delta \mathbf v_{\!\rm A} + \mathbf G_{\!\rm B} \cdot \delta \mathbf v_{\!\rm B} +
			\mathbf F \cdot \delta\mathbf v_{\rm C}
			\\&\quad
			- \mathbf F \cdot \delta\mathbf v_{\!\rm A} 
			+ \mathbf T \cdot \delta\bm \omega -  \mathbf T \cdot \delta\bm \Omega = \sum_{j=1}^{n_\sigma} \Pi_j\,\delta \sigma_{\!j}
		\end{split}
	\end{equation}%
	of the external active forces and torques,
	where the angular velocity of the pendulum is
	\begin{equation}\label{eq:pendangvel}
		\bm \Omega = \begin{bmatrix}
			\omega_{1}
			\\
			\frac{ 1}{h} (\sigma_{\gamma}{\!}-{\!} \omega_{2}{ R  \cos{\gamma}} )
			\\
			\omega_{3}
		\end{bmatrix}_{{\mathcal F}_{2}} .
	\end{equation}
	Thus, the pseudoforces become
	\begin{equation} \label{eq:pseudo_forces}
		\begin{split}
			\Pi_1 &= -F R + m g R \sin{\vartheta} - m_{1} g r \cos{\vartheta} 
			\\&\quad
			+ m_{2} g \big(R + h \cos{\gamma}\big) \sin{\vartheta}
			\,,\\ 
			\Pi_2 &= \frac{T}{h} \big({R \cos{\gamma}} + h\big) - m_{2} g R \sin{\gamma} \cos{\gamma} \cos{\vartheta} 
			\,,\\ 
			\Pi_3 &= - m_{2} g h \sin{\gamma} \sin{\vartheta}
			\,,\\ 
			\Pi_4 &= 	-F - m_{1} g \sin{\vartheta}
			\,,\\ 
			\Pi_5 &= -\frac{T}{h} + m_{2} g \sin{\gamma} \cos{\vartheta}\,.
		\end{split}
	\end{equation}
	
	The first set of equations in \eqref{eq:appell} are the dynamical equations, which assume the form \eqref{eq:dynamicEQS}. The second set of equations in \eqref{eq:appell} are the kinematic equations expressing the generalized velocities \eqref{eq:gen_coords} based on the pseudo-velocities \eqref{eq:pseudoVelocDefinition} as \eqref{eq:kinematicEQS}.

	The mass matrix in \eqref{eq:dynamicEQS} has the form 
	\begin{equation}\label{eq:massMtrx}
		\mathbf M = \begin{bmatrix}M_{11} & 0 & M_{13} & 0 & 0\\0 & M_{22} & M_{23} & 0 & 0\\M_{31} & M_{32} & M_{33} & 0 & 0\\0 & 0 & 0 & M_{44} & 0\\0 & 0 & 0 & 0 & M_{55}\end{bmatrix},
	\end{equation}
	where the matrix elements are
	\begin{equation}\label{eq:massMtrxElements}
		\begin{split}
			M_{11} &= {5 m R^{2}}/{4} + m_{1} r^{2} + m_{2} \big(R + h \cos{\gamma}\big)^{2},\\
			M_{13} &\equiv M_{31} = - m_{2} h \big(R + h \cos{\gamma}\big) \sin{\gamma},\\
			M_{22} &= R^{2} \big({3 m}/{2} + m_{1} + m_{2} \sin^{2}{\gamma}\big),\\
			M_{23} &\equiv M_{32} = - m_{1} R r,\\
			M_{33} &= {m R^{2}}/{4} + m_{2} h^{2} \sin^{2}{\gamma} + m_{1} r^{2},\\
			M_{44} &= m_{1}, \\
			M_{55} &= m_{2}.
		\end{split}
	\end{equation}
	The elements of the inertial forces $\mathbf C(\mathbf q, \bm\sigma)$ in \eqref{eq:dynamicEQS}, \eqref{eq:CandPI} are 
	{
	\setlength{\belowdisplayskip}{0pt}
	\setlength{\belowdisplayshortskip}{0pt} %
	\begin{align*}%
			C_1 & = \frac{1}{4}\Big(
			4 \omega_{1}^{2} m_{1} R  r + 8 \omega_{1}  \sigma_r m_{1} r 
			\\&
			+   \omega_{1} \omega_{2} m_{2} ((8 R^{2} \cos{\gamma} + 8 R h ) \sin{\gamma}- 8 R h \sin^{3}{\gamma})  
			\\&
			- \omega_{1} \sigma_{\gamma} m_{2}   ( 8 R + 8 h \cos{\gamma}) \sin{\gamma}  
			\\&
			+ \omega_{1} \omega_{3}  m_{2} (4 R h  + 4 h^{2} \cos{\gamma} ) \sin{\gamma} \tan{\vartheta}   
			\\&
			- \omega_{3} \sigma_{\gamma} m_{2} (8 R \cos{\gamma} + 8 h \cos^{2}{\gamma})  
			\\&
			+ \omega_{3}^{2} ((m R^{2} + 4 m_{1} r^{2} + m_{2} (4 R h \cos{\gamma} 
			\\&\quad+ 4 h^{2}  \cos^2{\gamma})) \tan{\vartheta})   
			\\&
			- \omega_{2} \omega_{3} (6 m R^{2} + 4 m_{1} R r \tan{\vartheta} 
			\\&
			\quad- m_{2} (4 R^{2} + 4 R h \cos{\gamma} )(\cos^2\gamma - \sin^2\gamma)) 
			\Big),
	\end{align*}}{
		\setlength{\abovedisplayskip}{0pt}
		\setlength{\abovedisplayshortskip}{0pt}
	\vspace{-1ex}\begin{equation}\label{eq:vecInertialForcesElements}
		\begin{split}
			C_2 &= \frac{1}{h}\Big(
			- \omega_{1}^{2} m_{2}  R h  ( R  + h \cos{\gamma}) \sin{\gamma} \cos{\gamma}
			\\&
			- \omega_{2}^{2} m_{2}  R^{3}  \sin{\gamma} \cos^{2}{\gamma}
			- \omega_{3}^{2}  m_{2} R h^{2} \sin^{3}{\gamma} 
			\\&
			- \sigma_{\gamma}^{2}  m_{2} R \sin{\gamma} 
			+ \omega_{1} \omega_{3} (m h R^{2}  -  m_{1} R^{2} h +  m_{2} R^{2} h \sin^{2}{\gamma} 
			\\&
			\quad + 2 m_{2} R h^{2} \sin^{2}{\gamma} \cos{\gamma} + m_{1} R h r \tan{\vartheta}) 
			\\&
			-\omega_{2} \omega_{3}  m_{2} R^{2} h  \sin{\gamma} \cos{\gamma} \tan{\vartheta} 
			\\&
			+ 2  \omega_{2} \sigma_{\gamma} m_{2} R^{2}  \sin{\gamma} \cos{\gamma} 
			- 2 \omega_{3} \sigma_r   m_{1} R h 
			\Big),
            \\
			C_3 &= \frac{1}{4}\Big(
			- 4 \omega_{3}^{2} m_{2} h^{2}  \sin{\gamma} \cos{\gamma} \tan{\vartheta} 
			\\&
			+ \omega_{1} \omega_{2} (2 m R^{2} - 8 m_{2} R h \sin^{2}{\gamma} \cos{\gamma}) 
			\\&
			+ \omega_{1} \omega_{3} (- m R^{2} \tan{\vartheta} + 4 m_{1} R r - 4 m_{2} h^{2} \sin^{2}{\gamma} \tan{\vartheta} 
			\\&\quad
			- 4 m_{1} r^{2} \tan{\vartheta}) 
			+ 8  \omega_{1} \sigma_{\gamma} m_{2} h  \sin^{2}{\gamma} 
			\\&
			+ \omega_{2} \omega_{3} (8 m_{2} R h \sin^{3}{\gamma} - 4 m_{2} R h \sin{\gamma})
			\\&
			+ 8 \omega_{3}  \sigma_r  m_{1} r 
			+ 8 \omega_{3} \sigma_{\gamma}   m_{2} h \sin{\gamma} \cos{\gamma} 
			\Big),
            \\
			C_4 &=  m_{1} R \omega_{2} \omega_{3} - m_{1} \omega_{1}^{2} r - m_{1} \omega_{3}^{2} r ,
			\\
			C_5 &= \frac{1}{h}\Big(
			\omega_{1}^{2}  (m_{2} R h \sin{\gamma} + m_{2} h^{2} \sin{\gamma} \cos{\gamma}) 
			\\&
			- \omega_{2}^{2}  m_{2} R^{2} \sin{\gamma} \cos{\gamma} 
			- \omega_{3}^{2} m_{2} h^{2}  \sin{\gamma} \cos{\gamma} 
			\\&
			+ \omega_{1} \omega_{3} (m_{2} R h \cos{\gamma} - 2 m_{2} h^{2} 	\sin^{2}{\gamma} + m_{2} h^{2}) 
			\\&
			+ \omega_{2} \sigma_{\gamma}  m_{2} R  \sin{\gamma} 
			\Big).
		\end{split}
	\end{equation}
	}

	\vspace{-10 mm}
	\section{Calculation of the contact force}\label{app:NewtonEuler_contactForce}

\begin{figure}[!t]
		\centering
		\includegraphics[scale=0.9]{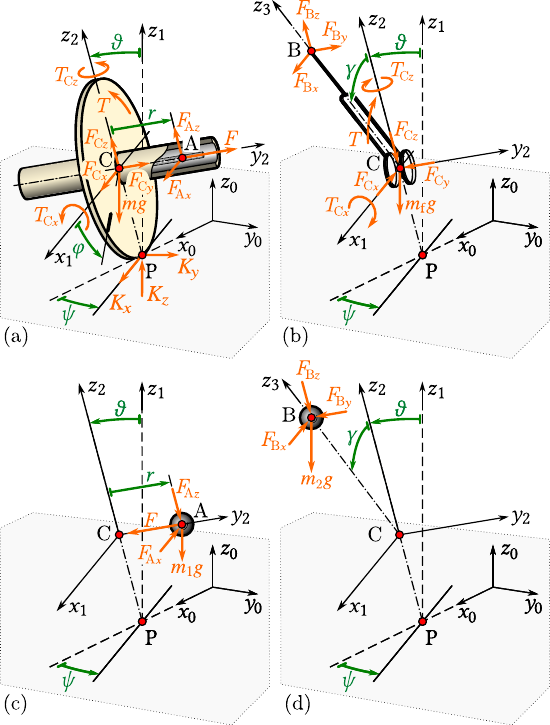}
		\caption{Free body diagrams of the parts.}
		\label{fig:freeBodyDiagrams}
	\end{figure}

	The Newton--Euler approach is used in order to calculate the constraining force $\mathbf K$ acting between the wheel and the ground. 
	The unicycle consists of a wheel, a (massless) fork, and two point masses, see Figure~\ref{fig:motivation};
	the corresponding free body diagrams are shown in Figure~\ref{fig:freeBodyDiagrams}.

	The Newton--Euler equations for the wheel are
	\begin{align}\label{eq:NEwheel1}
		m \mathbf{a}_{\mathrm C}  &= 
		m \mathbf{g} + \mathbf K + {\mathbf F}_\mathrm{A} + \bar{\mathbf F}_\mathrm{C} ,
		\\
		\label{eq:NEwheel2}
		\mathbf{J}_{\mathrm C} \bm \alpha 
		+ \bm{\omega}\times \mathbf{J}_{\mathrm C} \bm \omega 
		&=
		{\mathbf T}_\mathrm{C} 
		+ \mathbf{r}_{\mathrm{CP}}\times\mathbf K 
		+ \mathbf{r}_{\mathrm{CA}}\times {\mathbf F}_\mathrm{A}, 
		\intertext{%
			while for the fork they become
		}
		\label{eq:NEfork1}
			m_{\mathrm{f}} \mathbf{a}_{\mathrm C} &= m_{\mathrm{f}}\mathbf g + \bar{\mathbf F}_\mathrm{B} - \bar{\mathbf F}_\mathrm{C} ,
			\\
			\label{eq:NEfork2}
			\mathbf{J}_{\mathrm{f}} \bm\Lambda
			+ \bm{\Omega}\times \mathbf{J}_{\mathrm{f}} \bm \Omega 
			&=
			-{\mathbf T}_\mathrm{C} 
			+ \mathbf{r}_{\mathrm{CB}}\times\bar{\mathbf F}_\mathrm{B}.
		\intertext{%
			The Newton equation for point mass $m_1$ is
		}
		\label{eq:NEpm1}
		m_1 \mathbf{a}_{\mathrm A}  &= m_1 \mathbf{g} - {\mathbf F}_\mathrm{A} ,
		\intertext{%
			and for the point mass $m_2$ it is
		}
		\label{eq:NEpm2}
		m_2 \mathbf{a}_{\mathrm B}  &= m_2 \mathbf{g} - \bar{\mathbf F}_\mathrm{B} .
	\end{align}

	Equations \eqref{eq:NEwheel1}-\eqref{eq:NEpm2} form the Newton--Euler equations of the unicycle; 
	where the accelerations, angular velocities, and angular accelerations can be expressed as
    \begin{equation}
    \begin{split}
    &\mathbf a_{\mathrm{A}} \!= 
    \begin{bmatrix} a_{\mathrm{A}x} \\ a_{\mathrm{A}y} \\ a_{\mathrm{A}z} \end{bmatrix}_{{\mathcal F}_{2}}\!\!, \,\,\,
    \mathbf a_{\mathrm{B}} \!= 
    \begin{bmatrix} a_{\mathrm{B}x} \\ a_{\mathrm{B}y} \\ a_{\mathrm{B}z} \end{bmatrix}_{{\mathcal F}_{3}}\!\!, \,\,\,
    \mathbf a_{\rm C}  \!= 
    \begin{bmatrix} a_{\mathrm C x} \\ a_{\mathrm C y} \\ a_{\mathrm C z} \end{bmatrix}_{{\mathcal F}_{0}}\!\!, \,\,\,
    \mathbf{g} \!= 
    \begin{bmatrix} 0 \\ 0 \\ -g \end{bmatrix}_{{\mathcal F}_{0}}\!\!,
    \\
    &\bm\omega = 
    \begin{bmatrix} \omega_1 \\ \omega_2 \\ \omega_3 \end{bmatrix}_{{\mathcal F}_{2}}\!\!, \quad
    \bm\Omega  = 
    \begin{bmatrix} \Omega_{x} \\ \Omega_{y} \\ \Omega_{z} \end{bmatrix}_{{\mathcal F}_{2}}\!\!, \quad
    \bm\alpha = 
    \begin{bmatrix} \alpha_x \\ \alpha_y \\ \alpha_z \end{bmatrix}_{{\mathcal F}_{2}}\!\!, \quad
    \bm\Lambda = 
    \begin{bmatrix} \Lambda_{x} \\ \Lambda_{y} \\ \Lambda_{z} \end{bmatrix}_{{\mathcal F}_{2}}\!\!.
    \end{split}
    \end{equation}
	The internal force $ {\mathbf F}_\mathrm{A}$  and  torque $\mathbf{T}_\mathrm{C}$ consist of the actuator force $\mathbf F$  and torque $\mathbf{T}$ and the constraining force $\bar{\mathbf F}_\mathrm{A} $ and torque $\bar{{\mathbf T}}_\mathrm{C} $, respectively; these can be expressed with 
    \begin{equation}
    \begin{split}
    &{\mathbf F}_\mathrm{A} = \mathbf F + \bar{\mathbf F}_\mathrm{A},   \qquad\qquad\quad\,\,\,\,
    \mathbf{T}_\mathrm{C} = \mathbf{T} + \bar{{\mathbf T}}_\mathrm{C},
    \\
    &\mathbf F = 
    \begin{bmatrix} 0 \\ F \\0 \end{bmatrix}_{{\mathcal F}_{2}}\!\!, \,\,\,
    \bar{\mathbf F}_\mathrm{A} = 
    \begin{bmatrix} F_{\!\mathrm{A}x} \\ 0 \\ F_{\!\mathrm{A}z} \end{bmatrix}_{{\mathcal F}_{2}}\!\!, \,\,\,
    \mathbf T = 
    \begin{bmatrix} 0 \\ T \\ 0 \end{bmatrix}_{{\mathcal F}_{2}}\!\!, \,\,\,
    \bar{{\mathbf T}}_\mathrm{C} = 
    \begin{bmatrix} T_{\mathrm{C}x} \\ 0 \\ T_{\mathrm{C}z} \end{bmatrix}_{{\mathcal F}_{2}}\!\!.
    \end{split}
    \end{equation}
	The other constraining forces are the internal forces $\bar{\mathbf F}_\mathrm{B}, \bar{\mathbf F}_{\mathrm{C}}$ and the wheel-ground contact force $\mathbf K $, these are
    \begin{equation}
    \bar{\mathbf F}_{\mathrm{B}} = 
    \begin{bmatrix} F_{\!\mathrm{B}x} \\ F_{\!\mathrm{B}y} \\ F_{\!\mathrm{B}z} \end{bmatrix}_{{\mathcal F}_{2}}\!\!, \quad
    \bar{\mathbf F}_{\mathrm{C}} = 
    \begin{bmatrix} F_{\!\mathrm{C}x} \\ F_{\!\mathrm{C}y} \\ F_{\!\mathrm{C}z} \end{bmatrix}_{{\mathcal F}_{2}}\!\!, \quad
    \mathbf{K} = 
    \begin{bmatrix} K_x \\ K_y \\ K_z \end{bmatrix}_{{\mathcal F}_{1}}\!\!.
    \end{equation}
	The position vectors needed to calculate the torques of forces with respect to the center of masses are
    \begin{equation}\label{eq:geoMConstrS}
    \mathbf r_{\mathrm{CA}} = 
    \begin{bmatrix} 0 \\ r \\ 0 \end{bmatrix}_{{\mathcal F}_{2}}\!\!, \quad
    \mathbf r_{\mathrm{CB}} = 
    \begin{bmatrix} 0 \\ 0 \\ h \end{bmatrix}_{{\mathcal F}_{3}}\!\!, \quad
    \mathbf r_{\mathrm{CP}} = 
    \begin{bmatrix} 0 \\ 0 \\ -R \end{bmatrix}_{{\mathcal F}_{2}}\!\!.
    \end{equation}
	
	Considering frame $\mathcal F_0$, the components of \eqref{eq:NEwheel1} became
	\begin{align}\label{eq:NEwheel_comp}
			m a_{\mathrm C x} &=  (K_{x} + F_{\!\mathrm{A}x}+ F_{\!\mathrm{C}x})\cos{\psi} - K_{y} \sin{\psi},
			\nonumber\\&\quad
			- (F_{\!\mathrm{C}y} + F) \sin{\psi} \cos{\vartheta} 
			\nonumber\\&\quad
			+ (F_{\!\mathrm{A}z}  + F_{\!\mathrm{C}z}) \sin{\psi} \sin{\vartheta}
			\nonumber\\
			m a_{\mathrm C y} &=( K_{x} + F_{\!\mathrm{A}x}+ F_{\!\mathrm{C}x}  )\sin{\psi} + K_{y} \cos{\psi} ,
			\nonumber\\&\quad
			+(F_{\!\mathrm{C}y} + F )\cos{\psi} \cos{\vartheta}   
			\nonumber\\&\quad
			- (F_{\!\mathrm{A}z}  + F_{\!\mathrm{C}z} )\sin{\vartheta} \cos{\psi}
			\nonumber\\
			m a_{\mathrm C z} &= K_{z} - mg + (F_{\!\mathrm{A}z} + F_{\!\mathrm{C}z}) \cos{\vartheta} 
			\nonumber\\&\quad
			+(F_{\!\mathrm{C}y}   + F )\sin{\vartheta} ,
	\end{align}
	while the components of \eqref{eq:NEwheel2} in frame $\mathcal{F}_2$ are
	{\par\nobreak\vfil\penalty0\vfilneg\vtop\bgroup
	\begin{align}
		\tfrac{m R^{2}}{4} (\alpha_x - {\omega_{2} \omega_{3}})&= T_{\mathrm{C}x} + F_{\mathrm Az} r + K_{y} R \cos{\vartheta}
		\nonumber
		\\
		&\quad + K_{z} R \sin{\vartheta}\,,
		\nonumber
		\\
		\tfrac{m R^{2}}{2} \alpha_y &= T - K_{x} R\,,
		\\\nonumber
		\tfrac{m R^{2}}{4}(\alpha_z + { \omega_{1} \omega_{2}})&= T_{\mathrm{C}z} - F_{\mathrm Ax} r\,.
	\end{align}
	\par\xdef\tpd{\the\prevdepth}\egroup\prevdepth=\tpd}

	Neglecting the mass of the fork, ${ m_{\mathrm{f}}=0}$, the components of \eqref{eq:NEfork1} in frame  $\mathcal F_2$ are
	\begin{equation}
		\begin{split}
			0 &= F_{\!\mathrm{B}x} - F_{\!\mathrm{C}x},
			\\
			0 &= F_{\!\mathrm{B}y} - F_{\!\mathrm{C}y},
			\\
			0 &= F_{\!\mathrm{B}z} - F_{\!\mathrm{C}z}.
			\phantom{ - M -  F_{\!\mathrm{B}x} h \cos{\gamma} + F_{\!\mathrm{B}z} h \sin{\gamma},}
		\end{split}
	\end{equation}
	Similarly, neglecting the mass moment of inertia of the fork, 
	${\mathbf{J}_{\mathrm{f}} = \mathbf 0 }$, 
	\eqref{eq:NEfork2} resolved in frame  $\mathcal F_2$ yield
	\begin{equation}
		\begin{split}
			0 &= -  T_{\mathrm{C}x} - F_{\!\mathrm{B}y} h \cos{\gamma},
			\\
			0 &= - T +  F_{\!\mathrm{B}x} h \cos{\gamma} - F_{\!\mathrm{B}z} h \sin{\gamma},
			\\
			0 &= - T_{\mathrm{C}z} - F_{\!\mathrm{B}y} h \sin{\gamma}.
		\end{split}
	\end{equation}
	Equation \eqref{eq:NEpm1} resolved in frame $\mathcal F_2$ has the components
	\begin{equation}
		\begin{split}
			m_1 a_{\mathrm{A}x}&= - F_{\!\mathrm{A}x},
			\\
			m_1 a_{\mathrm{A}y}& = - F - m_{1} g \sin{\vartheta},
			\\
			m_1 a_{\mathrm{A}z}& = - F_{\!\mathrm{A}z} - m_{1} g \cos{\vartheta};
		\end{split}
	\end{equation}
	while \eqref{eq:NEpm2} in frame $\mathcal F_3$ gives 
	\begin{equation}\label{eq:NEpm2_comp}
		\begin{split}
			m_{2} a_{\mathrm B x}&= - F_{\!\mathrm{B}x} \cos{\gamma} + (F_{\!\mathrm{B}z} + m_{2} g \cos{\vartheta}) \sin{\gamma} ,
			\\
			m_{2} a_{\mathrm B y} &= -F_{\!\mathrm{B}y} - m_{2} g \sin{\vartheta} ,
			\\
			m_{2}  a_{\mathrm B z} &= - F_{\!\mathrm{B}x} \sin{\gamma} - (F_{\!\mathrm{B}z} + m_{2} g \cos{\vartheta}) \cos{\gamma} .
		\end{split}
	\end{equation}
	Then \eqref{eq:NEwheel_comp}--\eqref{eq:NEpm2_comp} form 18 equations for 25 unknowns which are the acceleration and constraint force/torque components:
	\begin{equation}\arraycolsep=3pt
		\begin{array}{llllllllll}
			a_{\mathrm A x},  & a_{\mathrm A y},   & a_{\mathrm A z}, & a_{\mathrm B x},   & a_{\mathrm B y},   & a_{\mathrm B z},  & a_{\mathrm C x},   & a_{\mathrm C y},   & a_{\mathrm C z},  \\
			\alpha_x,          & \alpha_y,          & \alpha_z,          & K_{x},            & K_{y},             & K_{z},           &
			F_{\!\mathrm{B}x}, & F_{\!\mathrm{B}y}, & F_{\!\mathrm{B}z}, \\
			F_{\!\mathrm{C}x}, & F_{\!\mathrm{C}y}, & F_{\!\mathrm{C}z}, & F_{\!\mathrm{A}x} & F_{\!\mathrm{A}z}, &  T_{\mathrm{C}x}, & T_{\mathrm{C}z}.   &
		\end{array}
	\end{equation}
	Due to the kinematic constraints \eqref{eq:kinemConstraints} and geometric constraints \eqref{eq:geom_constr_eq_z}, \eqref{eq:geoMConstrS}, the acceleration components are not independent.
	The acceleration components can be substituted from \eqref{eq:angularaccel_disc}--\eqref{eq:vB_aB}.
	Consequently, 18 independent unknowns remain which are
	\begin{equation}\arraycolsep=3pt
		\begin{array}{lllllllll}
			K_{x},             & K_{y},             & K_{z},           & F_{\!\mathrm{B}x}, & F_{\!\mathrm{B}y}, & F_{\!\mathrm{B}z}, & F_{\!\mathrm{C}x}, & F_{\!\mathrm{C}y}, & F_{\!\mathrm{C}z},  \\
			F_{\!\mathrm{A}x}, & F_{\!\mathrm{A}z}, & T_{\mathrm{C}x}, & T_{\mathrm{C}z}  & \dot\omega_{1},    & \dot\omega_{2},     & \dot\omega_{3},     & \dot\sigma_{r},    & \dot\sigma_{\gamma}.
		\end{array}
	\end{equation}
	Expressing the constraint forces and torques from \eqref{eq:NEwheel_comp}--\eqref{eq:NEpm2_comp} yields the components 
	\begin{equation}\label{eq:contactForceComps}
		\begin{split}
			K_x &=  \frac{m_2 \sin{\gamma} }{h}\Big(
			3 \omega_{2}\sigma_{\gamma}  R \cos{\gamma} - 2 \omega_{2}^{2}  R^{2} \cos^{2}{\gamma}  
			- \sigma_{\gamma}^{2} 
			\Big)
			\\&\quad 
			- m_{1} \big( \dot \omega_{3} r  + 2 \omega_{3} \sigma_r \big) 
			+ \dot \omega_{2} \big(
			m R 
			+ m_{1} R 
			+ m_{2} R \sin^{2}{\gamma}\big) 
			\\&\quad
			+ m_{2} \big(
			\dot \sigma_{\gamma} \cos{\gamma} 
			-  \omega_{2} \omega_{3} R \sin{\gamma} \cos{\gamma} \tan{\vartheta} 
			- \omega_{3}^{2} h  \sin{\gamma} 
			\big) 
			\\&\quad 
			+ \omega_{1} \omega_{3} \big(m R + m_{1} ( r \tan{\vartheta} - R ) 
			+ m_{2} (R + h \cos{\gamma})\big) 
			\,,\\
			K_y &=  
			\frac{m_2 \sin{\vartheta} }{h}\Big(
			\omega_{2} \sigma_{\gamma} R \big(3  \sin^{2}{\gamma}  
			- 2  \big) 
			+ \sigma_{\gamma}^{2}  \cos{\gamma}
			\\&\quad
			+ \omega_{2}^{2} R^{2}  \cos{\gamma} \big( 1 - 2  \sin^{2}{\gamma} 
			\big) 
			\Big)
			- \dot \omega_{1} \big(
			m R  + m_{1} r \tan{\vartheta} 
			\\&\quad%
			+ m_{2} ( R  + h \cos{\gamma} ) 
			\big) \cos{\vartheta}
			+ \omega_{1}^{2} \big(
			m R  \sin{\vartheta} 	
			\\&\quad%
			- m_{1} (R \sin{\vartheta}  + r \cos{\vartheta}) 
			+ m_{2} (R  + h  \cos{\gamma}) \sin{\vartheta}
			\big) 
			\\&\quad
			- \omega_{3}^{2} \big( 
			m_{1}  r ( { \sin{\vartheta} \tan{\vartheta}} + \cos{\vartheta}) +
			m_{2} h  \sin{\vartheta} \cos{\gamma} 
			\big)  
			\\&\quad
			+ \omega_{2} \omega_{3} \big(
			{m R}/{\cos{\vartheta}} + m_{1} R ({ \sin{\vartheta}}{\tan{\vartheta}} +  \cos{\vartheta}) 
			\\&\quad%
			+ m_{2} R (
			{ \cos^{2}{\gamma} \sin{\vartheta}} {\tan{\vartheta}} 
			-  (1-2  \sin^{2}{\gamma} ) \cos{\vartheta} 
			)
			\big) 
			\\&\quad
			+ m_{1} \big(\dot \sigma_r \cos{\vartheta}  - 2 \sigma_r \omega_{1} \sin{\vartheta} 
			\big) 
			+ m_{2} \big(
			\dot \sigma_{\gamma} \sin{\gamma} \sin{\vartheta} 
			\\&\quad%
			- \dot \omega_{2} R \sin{\gamma} \sin{\vartheta} \cos{\gamma} 
			+  \dot \omega_{3} h \sin{\gamma} \cos{\vartheta} 
			\\&\quad%
			- 2 \omega_{1} \omega_{2} R  \sin{\gamma} \cos{\gamma} \cos{\vartheta} 
			- 2 \omega_{1} \omega_{3} h \sin{\gamma} \sin{\vartheta} 
			\\&\quad%
			+ 2 \omega_{1} \sigma_{\gamma}  \sin{\gamma} \cos{\vartheta} 
			+ 2 \omega_{3} \sigma_{\gamma}  \cos{\gamma} \cos{\vartheta} 
			\big) 
			\,,\\
					K_z &= \frac{m_2 \cos{\vartheta}}{h}\Big(
			\omega_{2}^{2} R^{2} \big(2  \sin^{2}{\gamma}
			- 1\big)  \cos{\gamma}
			- \sigma_{\gamma}^{2} \cos{\gamma} 
			\\&\quad 
			+ \omega_{2} \sigma_{\gamma}  R \big(
			2  
			- 3  \sin^{2}{\gamma} \big) 
			\Big)
			- \dot \omega_{1} \big(
			m R \sin{\vartheta} 
			- m_{1} r \cos{\vartheta} 
			\\&\quad%
			+ m_{2} (
			R 
			+ h  \cos{\gamma}) \sin{\vartheta}
			\big) 
			-  \omega_{1}^{2} \big(
			m R  
			- m_{1} (R 
			- r \tan{\vartheta}) 
			\\&\quad%
			+ m_{2} (
			R 
			+ h \cos{\gamma} )
			\big) \cos{\vartheta}
			+ m_{1} \big( 
			\dot \sigma_r \sin{\vartheta}  
			+ 2 \sigma_r \omega_{1} \cos{\vartheta} 
			\big) 
			\\&\quad 
			+ m_{2} \big(
			\dot \omega_{2}  R \sin{\gamma} \cos{\gamma} \cos{\vartheta}
			+ \dot \omega_{3}h\sin{\gamma} \sin{\vartheta}
			\\
			&\quad%
			- \dot \sigma_{\gamma} \sin{\gamma} \cos{\vartheta}
			- {\omega_{3}^{2}h \sin{\vartheta} \cos{\gamma}}{\tan{\vartheta}}  
			\\
			&\quad%
			- 2  \omega_{1} \omega_{2} R \sin{\gamma} \sin{\vartheta} \cos{\gamma} 
			+ 2 \omega_{1} \sigma_{\gamma} \sin{\gamma} \sin{\vartheta} 
			\\&\quad%
			+  \omega_{1} \omega_{3}h(
			\cos{\vartheta}- { \sin{\vartheta}}{\tan{\vartheta}} 
			)  \sin{\gamma}
			+ 2 \omega_{3} \sigma_{\gamma} \sin{\vartheta} \cos{\gamma} 
			\\&\quad%
			+  \omega_{2} \omega_{3} R (3  \sin^{2}{\gamma} 
			- 2  )\sin{\vartheta}
			\big) 
			+ \big(m + m_{1} + m_{2}\big) g, 
		\end{split}
	\end{equation}
	of the contact force $\mathbf K$.
	Note that these contact force components depend on the control inputs $F$ and $T$ through the pseudo-accelerations according to \eqref{eq:dsigmaDynEQS}.

\printcredits

\bibliographystyle{cas-model2-names}

\end{document}